\documentclass[runningheads]{llncs}

\usepackage{eccv}

\usepackage{eccvabbrv}

\usepackage{graphicx}
\usepackage{booktabs}

\usepackage{pythonhighlight}
\usepackage[]{algorithm}
\usepackage{algorithmicx}
\usepackage{algpseudocode}
\usepackage{multirow}
\usepackage{wrapfig}
\usepackage{float}

\usepackage[accsupp]{axessibility}  

\usepackage[pagebackref,breaklinks,colorlinks,citecolor=eccvblue]{hyperref}

\usepackage{orcidlink}

\begin{document}

\title{Less is More: A Closer Look at Semantic-based Few-Shot Learning} 

\titlerunning{Abbreviated paper title}

\author{Chunpeng Zhou\inst{1} \and
Haishuai Wang*\inst{1} \and
Xilu Yuan\inst{1} \and
Sheng Zhou\inst{1} \and
Zhi Yu*\inst{1} \and
Jiajun Bu\inst{1} }

\authorrunning{F.~Author et al.}

\institute{Zhejiang University, China\\
\email{Corresponding Authors: \{haishuai.wang@gmail.com, yuzhirenzhe@zju.edu.cn\}}
}

\maketitle


\begin{abstract}

Few-shot Learning aims to learn and distinguish new categories from a scant number of images, presenting a significant challenge in the realm of deep learning. Recent researchers have sought to leverage the additional semantic or linguistic information of scarce  categories with a pre-trained language model to facilitate learning, thus partially alleviating the problem of insufficient supervision signals. Nonetheless, the full potential of the semantic information and pre-trained language model have been underestimated in the few-shot learning till now, resulting in limited performance enhancements. To address this,  we propose a straightforward and efficacious framework for few-shot learning tasks, specifically designed to exploit the semantic information and language model. More specifically, we explicitly harness the zero-shot capability of the pre-trained language model with learnable prompts. And we directly add the visual feature with the textual feature for inference without the intricate designed fusion modules as in prior  studies. Additionally, we apply the self-ensemble and distillation to further enhance performance. Extensive experiments conducted across four widely used few-shot datasets demonstrate that our simple framework achieves impressive results. Particularly noteworthy is its outstanding performance in the 1-shot learning task, surpassing the current state-of-the-art by an average of 3.3\% in classification accuracy. \footnote{We will make the source codes of the proposed framework publicly available upon acceptance.}
\end{abstract}

\section{Introduction}
\label{sec:intro}

Performing like humans is the ultimate goals of the Artificial Intelligent models. Recently, Deep learning-based technologies have made significant strides, achieving remarkable performances across various tasks, often rivaling or surpassing human capabilities in specific domains \cite{lecun2015deep,he2016deep,lin2017focal,He_2017_ICCV}. However, humans have the strong ability of Few-Shot Learning (FSL) \cite{fei2006one,lake2015human,wang2020generalizing}, which involves learning and discerning new classes with a very limited number of available samples. Despite the advancements in deep learning, FSL remains a significant challenge, showcasing a considerable performance gap between humans and deep learning models.  At the same time, studies in Human Neuroscience provide the compelling evidence about the hypothes that humans leverage both the visual and linguistic knowledge to comprehend novel concepts and categories \cite{jackendoff1987beyond,smith2005development,popham2021visual}. Inspired by these studies and aiming to alleviate the problem caused by limited visual supervision signals, a series of FSL research \cite{xing2019adaptive,peng2019few,liu2021cross,chen2023semantic} attempt to leverage the additional Semantic information of available samples  (e.g., textual information, also known as prompt) to assist models in recognizing rare classes, which imitate the human learning processes. 
For instance, AM3 \cite{xing2019adaptive} introduces an attention based fusion mechanism to fuse the visual and textual feature, guiding the positions of visual class prototypes. SP-CLIP \cite{chen2023semantic} proposes the semantic prompt, which utilizes the obtained textual semantic representations to guide the visual feature extraction network, employing two sophisticated complementary fusion mechanisms to integrate semantic representations into the feature extractor.

\begin{wrapfigure}[19]{r}{0.55\textwidth}
    \centering
    \vspace{-1cm}
    \subfloat[miniImageNet]{
		\includegraphics[width=3.2cm]{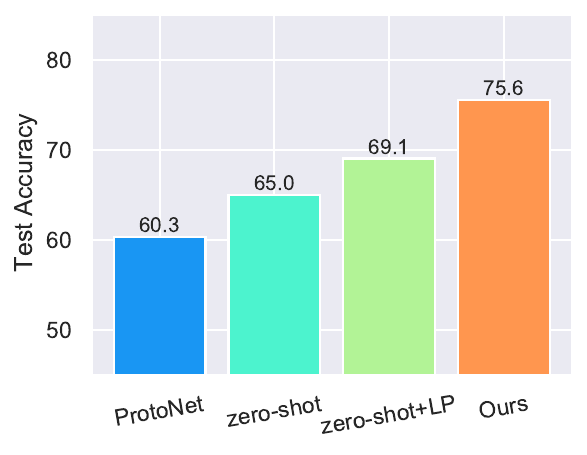}}
	\subfloat[CIFAR-FS]{
		\includegraphics[width=3.2cm]{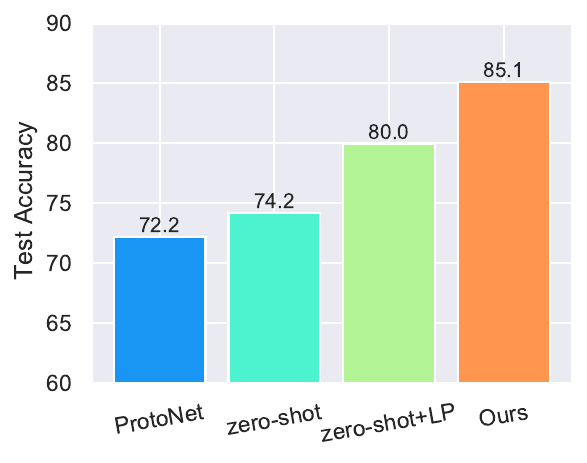}}
    \caption{5-way 1-shot Performance comparisons of simple few-shot image classification baselines with the pre-trained LM. The "zero-shot" align the visual feature and textual semantic feature, and do not use any samples from the novel classes.  We utilize the trained backbone to directly recognize novel classes. The "zero-shot+LP" denotes the learnable prompts (more details in Section 4) based on zero-shot. Two simple baselines outperform obviously than ProtoNet \cite{snell2017prototypical}, showing the distinguish  generalization capacity of LMs in FSL. }
    \label{fig:1}
\end{wrapfigure}

While these works \cite{xing2019adaptive,peng2019few,liu2021cross,yang2023semantic,chen2023semantic} partially alleviate the problem of insufficient supervision and achieve a certain performance improvements, they predominantly focus on how to design intricate multi-modal fusion modules to fuse the visual and semantic representations obtained by the visual encoder and textual encoder, respectively. However, these complex structures may potentially cause ignoring and even influence the generalization capacity of the used pre-trained language model (LM), leading to limited performance enhancements. At the same time, the recent research demonstrates the distinguished zero-shot capability of the pre-trained foundation language model \cite{brown2020language,radford2021learning,yao2023visual,openai2023gpt4}, which all trained on million-level or even billion-level language datasets. Consequently, we argue that these distinguished  generalization capability of the pre-trained LM should be considered, especially for the  scenarios with very limited supervision signals. To validate our motivation preliminarily, we design two simple few-shot image classification baselines with the pre-trained LM, and the experimental results have been depicted in Figure \ref{fig:1}. The "zero-shot" here means the we only use the base dateset to align the visual feature and textual semantic feature, without using any samples from the novel classes.  Subsequently, we employ the trained backbone with pre-trained LM to recognize novel classes directly without training. Following previous works \cite{radford2021learning,chen2023semantic}, the input prompt for the LM we chose is "A photo of a [classname]". We empirically observe that the simple zero-shot baseline outperforms obviously than the prototypical network \cite{snell2017prototypical} (ProtoNet in Figure \ref{fig:1}) in the setting of 5-way 1-shot learning both on miniImageNet and CIFAR dataset, despite the fact that ProtoNet can access the extra novel samples to help to recognize novel classes. Further, building on this simple zero-shot baseline, we adopt the learnable prompts instead of the pre-defined fixed prompts (the details about the learnable prompts can be found in Section 4) to further improve the generalization capacity \cite{zhou2022learning}, denoted with "zero-shot+LP" for short. We also observe the obvious performance improvements compared to the zero-shot baseline. The detail setting of these experiments can be found in the Appendix. These experimental results validate our motivation preliminarily and show the significant of the generalization capability of the pre-trained LM, which are usually ignored by previous FSL methods.  Inspired by this phenomenon, we aim to exploit the generalization capacity of the pre-trained LM in the FSL with very limited supervision signals. 


Consequently, we propose a straightforward and efficacious framework tailored for  few-shot learning tasks to maximize the utilization of the textual information and pre-trained LM in this paper. We aim to utilize the generalization capacity of the pre-trained LM directly to assist the classifier for few-shot classification, instead of designing the intricate and complex fusion modules as seen in previous works, which may ignore and even hurt the generalization capacity of the semantic features obtained by LMs.  In more details, our approach directly adds the visual feature obtained by a visual backbone and the textual semantic feature obtained by a pre-trained LM as the Multi-modal Feature Fusion mechanism, named as \textbf{SimpleFSL}. Although, there are lots of recent advanced multi-modal fusion mechanism \cite{gao2020survey}, we opt for the simplest Add operation in our proposed framework to validate our idea, and we argue this straightforward approach minimally impacts the generalization capability of the pre-trained LM. Despite having a simple network structure instead of the complex fusion mechanism, our proposed framework still attains satisfactory performances. Additional discussions on alternative fusion mechanisms are detailed in the experiments.
Addtionally, previous works \cite{xing2019adaptive,liu2021cross,chen2023semantic} utilize the fixed pre-defined prompts (e.g., a photo of a cat), which may constraining the generalization capability of the pre-trained LM \cite{zhou2022learning}.  To make a well adaptation of the pre-trained LM to diverse downstream datasets, we  adopt the more flexible learnable prompts as the input of LM, automating the prompt engineering, instead of the previous fixed prompt. Concretely, We model the context words of the prompt with learnable vectors, optimized in an end-to-end way via meta-training \cite{finn2017model}.  At the same time, we still keeping pre-trained LM frozen. In this way, we enable the FSL model to learn context-aware prompts autonomously, enhancing the flexible accommodation to various datasets without relying on handcrafted prompts. 
Furthermore, inspired by the recent advancements in Knowledge Distillation \cite{hinton2015distilling,zhang2018deep,zhang2019your}, we employ the self-ensemble and self-distillation mechanism based on the SimpleFSL to provide an additional boost for the FSL tasks, and we name this improved version as \textbf{SimpleFSL++}.

Our contributions can be summarized as following:
 \begin{enumerate}
	\item We explore a new perspective on Semantic-based FSL, and emphasize the significance of explicitly utilization of pre-trained language models for FSL.
	
	\item We introduce a novel and simple semantic-based few-shot learning framework, which exploits the pre-trained language model with learnable prompts via meta-learning. Further, we utilize the self-ensemble and self-Distillation to bring the additional performance improvement. 
	
	\item Extensive experiments across  four commonly used FSL benchmarks demonstrate the satisfactory performance of the proposed simple baselines compared to the state-of-the-art methods.

\end{enumerate}

%

\section{Related Work}
\subsection{Few-shot Learning}
In this paper, we focus on the few-shot learning task, which remains a very challenge topic in the realm of deep learning \cite{wang2020generalizing,song2023comprehensive}.  The inception FSL methods such as Prototypical Network (ProtoNet) \cite{snell2017prototypical}, Model-Agnostic Meta-Learning (MAML) \cite{finn2017model} utilize the meta-learning strategies, aiming at acquiring transferable features. Further, Meta-AdaM \cite{NEURIPS2023_ce26d216} proposes a mata-learned learning rate learner for more rapid convergence compared to MAML.  
Some recent advancements have pivoted towards leveraging relationships among available samples to enhance discriminative feature representation. For instance, GNNFSL \cite{garcia2017few} pioneers the utilization of the graph nerual network \cite{kipf2016semi} to explore the relations of samples. FEAT \cite{ye2020few} establishes class-wise relations via transformers \cite{vaswani2017attention} to derive more robust prototypical representations for inference. HGNN \cite{yu2022hybrid}  introduces a dual-graph neural network structure to exploit the relations of samples and classes, respectively.  Concurrently, another line of research involves pre-training effective feature extraction networks on base datasets \cite{chen2018closer, dhillon2019baseline, tian2020rethinking}, which then are transferred for inferring novel data. For instance, baseline++ \cite{chen2018closer} adopted the cosine similarity classifier to reduce intra-class variation among features during training. RFS \cite{tian2020rethinking} leverages the Born-again strategy \cite{furlanello2018born} to enhance pre-training. SUN \cite{dong2022self} utilizes the individual supervision for local semantic learning, which helps to learn generalizable patterns in FSL.

\subsection{Semantic-based Few-shot Learning}
Recent endeavors in FSL have incorporated auxiliary semantic  information and pre-trained language models to assit the recognition of novel classes \cite{xing2019adaptive,peng2019few,liu2021cross,yang2023semantic,chen2023semantic}. For instance, AM3 \cite{xing2019adaptive} introduces an attention based fusion mechanism to integrate the visual and textual features, guiding the learning of class prototypes.  CMGNN \cite{liu2021cross} proposes a Cross-Modality Graph Neural Network to generate meta nodes with semantic information, which aids the corresponding visual feature learning. SP-CLIP \cite{chen2023semantic} proposed the semantic prompts, which utlizes the obtained textual semantic representations to guide the the visual feature extraction network, employing two complementary fusion mechanisms to insert semantic representations into the feature extractor. As discussed above, although these semantic-based approaches have shown performance gains, they all predominantly concentrate on intricate fusion modules to leverage visual and textual information, inadvertently overlooking the potential of pre-trained language models. Consequently, we explore a straightforward FSL framework to explicitly exploit the generalization capability of the pre-trained language model with learnable prompts.

\section{Preliminary}
\subsection{Problem formulation}


In the context of Few-Shot Learning (FSL), the objective of a model is to recognize unknown samples from unseen or novel classes via leveraging a very limited number of available samples. Formally, the dataset encompassing novel classes, is denoted as $\mathbf{D_{novel}}$.  We follow the previous N-way K-shot learning setting \cite{vinyals2016matching,snell2017prototypical,finn2017model}, where an FSL task comprises N classes with K labeled samples for per class, and these labeled samples are named as the support set in novel classes.  Conversely, the unknown or unlabeled samples in novel classes, which the model aims to classify, are denoted as the query set in each FSL task.

\subsection{Meta-training}
Training an FSL model directly from scratch only with the limited number of labeled samples (i.e., one per class) poses significant difficulties and may cause a high risk of overfitting. Consequently, an additional base dataset with all samples annotated, denoted as $\mathbf{D_{base}}$, is provided to pre-train the FSL model to alleviate overfitting in the training phase \cite{vinyals2016matching,snell2017prototypical,chen2018closer}. In order to reduce the gap between the full labeled $\mathbf{D_{base}}$ and very few partial labeled $\mathbf{D_{novel}}$, prevalent FSL works usually adopt the meta-training strategy \cite{finn2017model}, a.k.a. episodic training. In details, meta-training endeavors to sample a series of N-way K-shot learning tasks per episode from $\mathbf{D_{base}}$. Each episode also contains a support set and a query set sampled from $\mathbf{D_{base} }$, as have been done from $\mathbf{D_{novel}}$. The FSL model will be trained with $\mathbf{D_{base}}$ via meta-training until the FSL model converges. It is crucial to note that the class categories in $\mathbf{D_{base} }$ and $\mathbf{D_{novel} }$ are entirely disjoint, without any overlap in the their label spaces or samples, thus ensuring the model's capacity for generalizing to unseen classes with limited labeled data.

\section{Method}

\begin{figure*}[t]
	\centering
	\includegraphics[width=12cm]{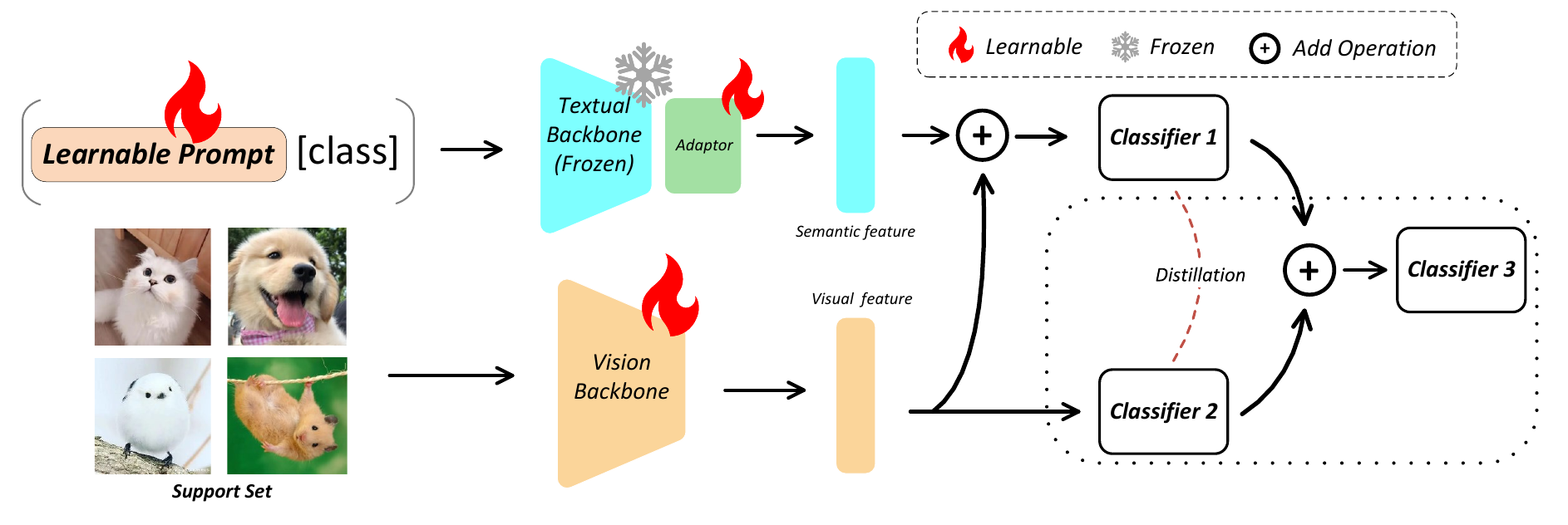}
        \vspace{-0.2cm}
	\caption{The schematic of our proposed SimpleFSL and SimpleFSL++. Instead of designing the intricate fusion modules as seen in previous works, we directly add the obtained visual feature by a visual backbone with the obtained textual semantic feature for few-shot classification tasks.  We also adopt the more flexible learnable prompts as input to the LM to automate prompt engineering, instead of the previous fixed prompts. We name this method as SimpleFSL, and we further utilize the self-ensemble and self-distillation to further improve performance (the dotted box part), encapsulated in SimpleFSL++.}
	\label{fig:2}
	\vspace{-0.5cm}
\end{figure*}

In this section, we introduce the details of our proposed framework SimpleFSL and its variants SimpleFSL++, as both depicted in Fig.\ref{fig:2}. The whole framework is very simple without complex architectures, which contains three primary modules: a visual backbone for visual feature extraction from the images, a textual backbone with input prompts for the textual semantic feature extraction, and feature fusion module for the final prediction. Additionally, SimpleFSL++ contains the self-ensemble and self-distillation module to improve the performance further.

\subsection{Pre-training}

Following previous works \cite{ye2020few,chen2021meta,chen2023semantic}, we first pre-train our visual backbone on  $\mathbf{D_{base}}$ to expedite convergence, before the meta-training. 
During the pre-training stage, the visual features of all samples in $\mathbf{D_{base}}$ are extracted by a visual backbone $f_\Theta (\ )$ (e.g., Visual Transformer \cite{dosovitskiy2020image}) with learnable parameters $\Theta$. Subsequently, we use a simple linear classifier with learnable parameters $\mathbf{\{W,b\}}$ comprising a weight term $\mathbf{W}$ and a bias term $\mathbf{b}$, which maps the input features into one of the base classes. This procedure is optimized by minimizing the standard cross entropy loss, described formally as follows:
\begin{equation}\small
	\mathcal{L}_{\text {pre }}=\frac{1}{\left|D_{\text {base }}\right|} \sum_{(\boldsymbol{x}, y) \in D_{\text {base }}}-\log \frac{\exp \left(\boldsymbol{W}_y^T f(\boldsymbol{x})+\boldsymbol{b}_y\right)}{\sum_i \exp \left(\boldsymbol{W}_i^T f(\boldsymbol{x})+\boldsymbol{b}_i\right)}  +\mathcal{R}
\end{equation}
where $\boldsymbol{x}$ represents an image in $\mathbf{D_{base}}$, $y$ is  the corresponding ground-truth label, and $\mathcal{R} $ denotes the L-2 regularization term.

\subsection{Context-aware Prompt}
As illustrated in the preceding section, the visual feature can be obtained by a visual backbone for the downstream tasks. 
Notwithstanding, reliance solely on the visual information may cause insufficient supervision signals and suboptimal performance \cite{xing2019adaptive}.
To supplement more available input information, previous works explore additional semantic information as the prompts to guide the visual feature extraction. Formally, given an image $\boldsymbol{x}$ from the support set, they pre-define a fixed prompt using the class name of $\boldsymbol{x}$, e.g. \textit{cat} \cite{xing2019adaptive,liu2021cross} or \textit{a photo of a cat} \cite{chen2023semantic}. The pre-defined prompt, denoted as $y^{text}$, is related with the class label. Then, the textual semantic feature will be procured by an off-the-shelf pre-trained language model (e.g., CLIP \cite{radford2021learning}), denoted as $g(\ )$, formulated as:
\begin{equation}
	g(y^{text}) = g( \text{a photo of [class nanme]} )
\end{equation}

However, these pre-defined fixed prompts may not be flexible enough for various downstream tasks and previous works found that this may lead to the inferior performance \cite{zhou2022learning}. Consequently, we address this by modeling the input prompts with continuous learnable parameters, which will be optimized in the end-to-end learning way, thereby allowing for flexible accommodation to various datasets. Specifically, the text prompt $y^{text}$ is instantiated as a learnable vector $\mathbf{v}$ combined with the class name and shared with all classes in our implements. The learnable prompt  $y^{text}_v$ fed into the text encoder $g(\ )$ takes the form:
\begin{equation}
	g(y^{text}_\mathbf{v}) = g( v_1,v_2,  \ldots  ,v_L, \text{[class name]} )
\end{equation}
where $v_i$ is the $i$-th term of the learnable prompt $\mathbf{v}$, and $L$ denotes the length of context tokens. Specially, the dimension $d_{text}$ of prompt $\mathbf{v}$ keep the same with the language model. In our method, we set $L = 4$ and $d_{text} = 512$ across all experimental setups.

\subsection{Multi-modal Feature Fusion}
Thus far, we have extracted  the visual feature $f(x)$ from a visual backbone and the textual semantic feature  $g(y^{text}_\mathbf{v})$ from a pre-trained LM. Previous works \cite{chen2023semantic, liu2021cross} have employed  the intricate designed fusion modules to utilize visual and textual information simultaneously. However, these complex fusion methods may inadvertently compromise the generalization capability of the pre-trained LM. Consequently , the potential of semantic information for FSL has been underestimated.  
To address this, we straightforwardly add  the visual feature $f(x)$ and the textual semantic feature  $g(y^{text}_\mathbf{v})$ as the Multi-modal Feature Fusion, and we name this simple baseline as \textbf{SimpleFSL}. Despite the availability of advanced multi-modal fusion techniques \cite{gao2020survey},   we choose the simplest Add operation in our proposed framework to demonstrate our concept. We argue this simple add operation has little effect with the generalization capability of the pre-trained LM. The more discussions about the alternative multi-modal fusion operations can be found in the experiments section.

While there is the inherent discrepancy between visual and textual modalities, including differing dimensions. Following previous works \cite{chen2023semantic,liu2021cross,xing2019adaptive}, we adopt an adaptor here to transform the semantic feature space into the visual feature space, and ensure dimensional consistency. We choose a simple two-layer Multi-Layer Perceptron (MLP) with the non-linear activation function as the adaptor in our implements. The discussion about other adaptors can be found in the experiments section. The transformation  process is expressed as:
\begin{align}
	\boldsymbol{z} &= adaptor(g(y^{text}_\mathbf{v})) 
	  = \boldsymbol{W_2} \  \sigma ( \boldsymbol{W_1} \ g(y^{text}_\mathbf{v}) + \boldsymbol{b_1} ) + \boldsymbol{b_2}
\end{align}
where $\boldsymbol{W_1}$, $\boldsymbol{W_2}$, $\boldsymbol{b_1}$, and $\boldsymbol{b_2}$ are the learnable parameters of the adaptor module, $\sigma$ is the non-linear activation function. 

Subsequently, we compute the class prototypes \cite{snell2017prototypical} by averaging the sum of visual features $f(\boldsymbol{x})$ and the transformed semantic feature  $\boldsymbol{z}$ for each sample in the support set, formulated as:
\begin{align}
	\boldsymbol{p_i} & = \frac{1}{K}  \sum_{j=1}^{K}  ( f(\boldsymbol{x_j}) + \boldsymbol{z_j} )
\end{align}
where $\boldsymbol{p_i}$ denotes the obtained prototype of $i$-th class.  Classification of an unlabeled sample $\boldsymbol{x_q}$ from the query set is performed using a non-parametric distance-based classifier \cite{chen2021meta,snell2017prototypical} with a softmax function: 
\begin{equation}
	\hat{y}_q =\frac{\exp \left(\left\langle f(\boldsymbol{x_q}), \boldsymbol{p_i} \right\rangle         \right)}{\sum_j \exp \left(\left\langle f(\boldsymbol{x_q}), \boldsymbol{p_j} \right\rangle    \right)}
\end{equation}
where $\langle \ ,\ \rangle$ denotes the cosine similarity of two vectors, and $\hat{y}_q$ is the predicted label for the query sample $\boldsymbol{x_q}$. During meta-training, we freeze the pre-trained LM $g(\ )$ and only update the other learnable parameters by minimizing the cross-entropy loss: $L_{1} = \sum_{D_{base}} CE(\hat{y}_q ,y_q)$, 
where $CE( \ )$ represents the cross-entropy function.

\subsection{Self-ensemble and Self-Distillation}
Building upon SimpleFSL, we introduce \textbf{SimpleFSL++}, which incorporates self-ensemble and self-distillation modules to further refine FSL performance.  Recall that we have obtained the fusion feature for inference, and we also have obtained the visual feature $f(\boldsymbol{x})$ from the visual backbone, which can also serve as the input of a classifier, as shown in Fig.\ref{fig:2}. Similarly, we have: 
$\boldsymbol{p_i^{0}} = \frac{1}{K}  \sum_{j=1}^{K}  f(\boldsymbol{x_j})$ and $\hat{y}_q^{0} =\frac{\exp \left(\left\langle f(\boldsymbol{x_q}), \boldsymbol{p_i^{0}} \right\rangle     / \tau    \right)}{\sum_j \exp \left(\left\langle f(\boldsymbol{x_q}), \boldsymbol{p_j^{0}} \right\rangle  / \tau  \right)}$.

Inspired the research in Knowledge Distillation \cite{hinton2015distilling,zhang2018deep,zhang2019your}, we utilize the self-ensemble mechanism to build the classifier-3, which ensembles the prediction of $\hat{y}_q^{0}$ and $\hat{y}_q$, obtained from classifier-1 and classifier-2 respectively, as shown in \Cref{fig:2}. In this way, the prediction will be more robust and improved further. Mathematically, the predictions of SimpleFSL++ is written as:
\begin{equation}\label{eq:11}
	\hat{y}_q^{++} =  \hat{y}_q + \lambda  \hat{y}_q^{0}
\end{equation}
where  $\lambda$ is the weighting factor to balance these two classifiers. Additionally, we further utilize the self-Distillation mechanism to allow these classifiers learn  reciprocally, which can transfer the learned knowledge and serve as the regularization mutually \cite{yuan2020revisiting,zhang2018deep}.  In detail, the self-distillation mechanism employed is formulated as:
\begin{equation}
	L_{KD} = \sum_{D_{base}} \frac{1}{2} ( KL(\hat{y}_q,\hat{y}_q^{0})   +  KL(\hat{y}_q^{0},\hat{y}_q) )
\end{equation}
where $KL$ refers to the knowledge distillation loss, and we opt Kullback-Leibler Divergence as our implementation.

To summarize,  the comprehensive loss of the proposed SimpleFSL++ is articulated as:
\begin{equation}\label{eq:13}
	L^{++} = L_1  + L_2 + \alpha L_{KD}
\end{equation}
where $L_{2} = \sum_{D_{base}} CE(\hat{y}_q^{0} ,y_q)$ represents the cross-entropy loss for the classifier-2, and  $\alpha$ is a hyper-parameters. 
Additionally, it is worth to note that self-distillation is not required during inference.

\section{Experiments}

\subsection{ Datasets and Implementation details.}
We conduct the FSL experiments on four widely used datasets: miniImageNet \cite{vinyals2016matching}, tieredImageNet \cite{ren2018meta}, CIFAR-FS \cite{dhillon2019baseline,krizhevsky2009learning} and FC100 \cite{oreshkin2018tadam}.  The first two are subsets of ILSVRC-12 dataset \cite{russakovsky2015imagenet}, and the last two derive from the CIFAR-100 dataset \cite{krizhevsky2009learning}. These datasets are all public available. 
\textbf{MiniImageNet}:  This dataset comprises 100 classes and 60,000 images, in which 64 classes are assigned to the base dataset, 16 classes to the validation dataset, and 20 classes to the novel dataset. 
\textbf{TieredImageNet}: It is a larger dataset compared to MiniImageNet, containing 608 classes and 779,165 images, and the base and novel datasets of it is more semantically different. We follow the previous split proposed by \cite{ren2018meta}, in which 351, 97 and 160 classes are used for the base, validation, and novel datasets, respectively. 
\textbf{CIFAR-FS}: It contains 100 classes and 600 images per class. Following previous works \cite{dhillon2019baseline}, we use 64 classes for the base dataset, 16 classes for the validation dataset, and the remaining 20 classes for the novel dataset.
\textbf{FC100}: It also contains 60,000 images , with 100 classes, and was split into 60 base classes, 20 validation classes and 20 novel classes, according their semantic superclasses. Consequently, its discernible semantic gap presents a steeper challenge compared to CIFAR-FS.

\begin{table}[tb] 
\caption{5-way 1/5-shot classification accuracy (\%) and 95\% confidence interval on miniImageNet and tieredImageNet. Methods in the top of first rows do not use semantic informations, and methods in the middle rows leverage the semantic informations. $\dag$ means we re-implement the experiments with its open code. }
	\centering

\resizebox{0.95\textwidth}{!}{                
	\begin{tabular}{lccccccc}
        
		\hline
		\multirow{2}{*}{Methods} &   \multirow{2}{*}{Backbone} & \multirow{2}{*}{Params/FLOPs}& \multicolumn{2}{c}{miniImageNet} & \multicolumn{2}{c}{tieredImageNet} \\ 
		\cline{4-7} &&& 5-way 1-shot &\multicolumn{1}{c}{5-way 5-shot} &5-way 1-shot & \multicolumn{1}{c}{5-way 5-shot} \\ \hline
		
		 ProtoNet \cite{snell2017prototypical} &    ResNet-12 & 12.5M/$3.5 \times 10^9$  & $60.34 \pm 1.20$ & $80.54 \pm 1.13$ & $69.63 \pm 0.53$ & $84.82 \pm 0.36$\\
		 MAML  \cite{finn2017model} &  ResNet-12 & 12.5M/$3.5 \times 10^9$& $58.05 \pm 0.10$ & $72.41 \pm 0.20$ & $63.85 \pm 0.76$ & $81.57 \pm 0.56$\\
		Fine-tuning \cite{dhillon2019baseline} &  Wide-ResNet &  36.5M/$3.7 \times 10^{10}$ & $57.73 \pm 0.62 $& $78.17 \pm 0.49$ & $66.58 \pm 0.70$ & $85.55 \pm 0.48$ \\
		 FEAT  \cite{ye2020few} &  ResNet-12 & 12.5M/$3.5 \times 10^9$& $66.78 \pm 0.20$ & $82.05 \pm 0.14$ & $66.78 \pm 0.20$ & $82.05 \pm 0.14$\\
		 Neg-Margin \cite{liu2020negative} & ResNet-12 & 12.5M/$3.5 \times 10^9$& $63.85 \pm 0.76$ & $81.57 \pm 0.56$ & $63.85 \pm 0.76$ & $81.57 \pm 0.56$\\
		RFS \cite{tian2020rethinking} &    ResNet-12 & 12.5M/$3.5 \times 10^9$& $62.02 \pm 0.63$ & $79.64 \pm 0.44$ & $71.52 \pm 0.69$ & $86.03 \pm 0.49$ \\
		
		Align  \cite{afrasiyabi2020associative} &    Wide-ResNet & 36.5M/$3.7 \times 10^{10}$& $65.92 \pm 0.60$ & $82.85 \pm 0.55$ & $74.40 \pm 0.68$ & $86.61 \pm 0.59$\\
		
		 FRN  \cite{wertheimer2021few} &    ResNet-12 & 12.5M/$3.5 \times 10^9$& $66.45 \pm 0.19$ & $82.83 \pm 0.13$ & $71.16 \pm 0.22$ & $86.01 \pm 0.15$\\
		 MixtFSL \cite{afrasiyabi2021mixture} &  ResNet-12 & 12.5M/$3.5 \times 10^9$& $63.98 \pm 0.79$ & $82.04 \pm 0.49$ & $70.97 \pm 1.03$ & $86.16 \pm 0.67$\\
		 MixtFSL  \cite{afrasiyabi2021mixture} &  Wide-ResNet& 36.5M/$3.7 \times 10^{10}$ & $64.31 \pm 0.79$ & $81.66 \pm 0.60$ & — & — \\
		HGNN \cite{yu2022hybrid} &   ResNet-12 & 12.5M/$3.5 \times 10^9$& $67.02 \pm 0.20$ & $83.00 \pm 0.13$ & $72.05 \pm 0.23$ & $86.49 \pm 0.15$ \\
		
		MTL \cite{wang2021bridging} & ResNet-12 &  $12.5 \mathrm{M} / 3.5 \times 10^9$ & $59.84 \pm 0.22$ & $77.72 \pm 0.09$ & $67.11 \pm 0.12$ & $83.69 \pm 0.02$ \\
		
		 SetFeat \cite{afrasiyabi2022matching} & ResNet-12 & 12.5M/$3.5 \times 10^9$ & $68 . 32 \pm 0 . 62$ & $8 2 . 71 \pm 0 . 46$ &  $7 3 . 63 \pm 0. 88$ & $87 . 59 \pm 0 . 57$ \\
         Pre-train \cite{chen2023semantic} &   Visformer-T  & 10.0M/$1.3 \times 10^9$& $65.16 \pm 0.44$ & $81.22 \pm 0.32$ & $72.38 \pm 0.50$ & $86.74 \pm 0.34$ \\
		 SUN \cite{dong2022self} & Visformer-S & $12.4 \mathrm{M} / 1.7 \times 10^8$ & $67 . 80 \pm 0 . 4 5$ & $8 3 . 25 \pm 0 . 3 2$ &  $7 2 . 00 \pm 0. 50$ & $86 . 74 \pm 0 . 3 3$ \\
   Meta-AdaM \cite{NEURIPS2023_ce26d216} & ResNet-12 & 12.5M/$3.5 \times 10^9$ & $59 . 89 \pm 0 . 49$ & $77 . 92 \pm 0 . 4 3$ & $65 . 31 \pm 0 . 48$ &  $85 . 24 \pm 0 . 3 5$ \\
   CORL \cite{He_2023_WACV} & ResNet-12 & 12.5M/$3.5 \times 10^9$ & $65 . 74 \pm 0 . 53$ & $83 . 03 \pm 0 . 3 3$ & $73 . 82 \pm 0 . 58$ &  $86. 76 \pm 0 .53$ \\
		 
		\hline 
		
		KTN \cite{peng2019few} &     ResNet-12 & 12.5M/$3.5 \times 10^9$& $6 1 . 42 \pm 0. 72$ & $74 . 16 \pm 0 . 56$ & — & — \\
		AM3 \cite{xing2019adaptive} &     ResNet-12 & 12.5M/$3.5 \times 10^9$& $6 5 . 30 \pm 0. 49$ & $78 . 10 \pm 0 . 36$ & $69 . 08 \pm 0. 47$ & $8 2 . 5 8 \pm 0 . 31$\\
		TRAML \cite{li2020boosting} &     ResNet-12 & 12.5M/$3.5 \times 10^9$& $67.10\pm0.52$ & $79.54\pm0.60$ &  — &  —\\
		DeepEMD-BERT \cite{yan2021aligning} &     ResNet-12 & 12.5M/$3.5 \times 10^9$& $67.03\pm0.79$ & $83.68\pm0.65$ & $73.76\pm0.72$ &  $87.51\pm0.75$ \\
		CMGNN-DPGN \cite{liu2021cross} &     ResNet-12 & 12.5M/$3.5 \times 10^9$& $71.38\pm0.51$ & $ 82.60\pm0.47$ & $72.89\pm0.49$ &  $84.92\pm0.48$ \\
        LEP-CLIP \cite{yang2023semantic} &     ResNet-12 & 12.5M/$3.5 \times 10^9$& $71.64\pm0.40$ & $79.67\pm0.32$ & $73.88\pm0.48$ &  $84.88\pm0.36$ \\
		SP-CLIP $^{\dag}$ \cite{chen2023semantic} &     Visformer-T & 10.0M/$1.3 \times 10^9$& $72.41\pm0.40$ & $83.23\pm0.65$ & $77.83\pm0.87$ &  $87.56\pm0.64$ \\
		
		\hline 
		\textbf{SimpleFSL} (Ours)  &     Visformer-T & 10.0M/$1.3 \times 10^9$& $\mathbf{74.80} \pm \mathbf{0. 66}$ & $\mathbf{8 3 . 34} \pm \mathbf{0 . 5 5}$ & $\mathbf{80 . 06} \pm \mathbf{0. 81}$ & $\mathbf{8 8 . 33} \pm \mathbf{0 . 62}$ \\

		\textbf{SimpleFSL++} (Ours) &     Visformer-T & 10.0M/$1.3 \times 10^9$ & $\mathbf{75.59} \pm \mathbf{0. 37}$ & $\mathbf{8 3 . 89} \pm \mathbf{0 . 5 4}$ & $\mathbf{80 . 52} \pm \mathbf{0. 81}$ & $\mathbf{8 8 . 36} \pm \mathbf{0 . 60}$ \\
		\hline 
	\end{tabular}
	}

 \label{tab:1}
\end{table}

Our proposed SimpleFSL and SimpleFSL++ both contain the visual backbone and textual backbone for the feature extraction. 
we opt the Visformer-Tiny \cite{chen2021visformer} as the visual backbone, and resize the all input image with 224×224 pixel. Compared to the usually used ResNet-12 \cite{he2016deep}, the Visformer-Tiny has the similar number of parameters but with less floating point operations (FLOPS). And unless otherwise specified, the dimension of obtained visual representation is 384. 
we adopt the text encoder of pre-trained CLIP \cite{radford2021learning} as the textual backbone, due to the fact that it has distinguish performance \cite{chen2023semantic} and supports the learnable prompt tuning, as described above.  The code and pre-trained weights of CLIP are available for public use \footnote{https://github.com/openai/CLIP.}. Notably, we do not use the visual encoder of CLIP for a fair comparison. During the training stage, we freeze the all parameters of textual backbone and optimize only the other model parameters. We adopt the embeddings of “a photo of a” to initialize the learnable prompt $\mathbf{v}$, and the dimension of obtained textual representation is consistently 512.

In pre-training and meta-training stages, we all employ the AdamW optimer \cite{loshchilov2018decoupled} with a learning rate of 5e-4 and a weight decay of 5e-2. Specially, we reduce the learning rate of the visual backbone to 1e-6 during the meta-training stage, keeping others the same. For evaluation, we test our framework under 5 way-1 shot/5 shot settings on the novel dataset and randomly sample 2,000 few-shot tasks from it. Then, we report the top-1 mean accuracy (\%) with the 95\% confidence interval.  Notably, our proposed frameworks both do not require fine-tuning during evaluation, while some FSL baselines require \cite{afrasiyabi2021mixture,dhillon2019baseline,finn2017model}. All  experiments are conducted on a Linux machine with a single NVIDIA RTX3090 GPU.

\subsection{Main results}
\label{sec:52}

\begin{table*}[tb] 
\caption{5-way 1/5-shot classification accuracy (\%) and 95\% confidence interval on  CIFAR-FS and FC100. }
	\centering
\resizebox{0.95\textwidth}{!}{
\begin{tabular}{lcccccc}
		\hline
	\multirow{2}{*}{Methods} &   \multirow{2}{*}{Backbone} & \multirow{2}{*}{Params/FLOPs}& \multicolumn{2}{c}{CIFAR-FS} & \multicolumn{2}{c}{FC100 } \\ 
	\cline{4-7} &&& 5-way 1-shot &\multicolumn{1}{c}{5-way 5-shot} &5-way 1-shot & \multicolumn{1}{c}{5-way 5-shot} \\ \hline
	
	\hline Self-Supervised [14] & WRN-28-10 & $36.5 \mathrm{M} / 3.7 \times 10^{10}$ & $69.55 \pm 0.34$ & $82.34 \pm 0.24$ & - & - \\
	Align \cite{afrasiyabi2020associative} & WRN-28-10 & $36.5 \mathrm{M} / 3.7 \times 10^{10}$ & - & - & $4 5 . 8 3 \pm 0 . 4 8$ & $59.74 \pm 0.56$ \\
	ProtoNet \cite{snell2017prototypical} & ResNet-12 & $12.5 \mathrm{M} / 3.5 \times 10^9$ & $72.2 \pm 0.7$ & $83.5 \pm 0.5$ & $37.5 \pm 0.6$ & $52.5 \pm 0.6$ \\
	MetaOptNet \cite{lee_meta-learning_2019} & ResNet-12 & $12.5 \mathrm{M} / 3.5 \times 10^9$ & $72.6 \pm 0.7$ & $84.3 \pm 0.5$ & $41.1 \pm 0.6$ & $55.5 \pm 0.6$ \\
	MABAS \cite{kim_model-agnostic_2020} & ResNet-12 & $12.5 \mathrm{M} / 3.5 \times 10^9$ & $73.51 \pm 0.92$ & $85.49 \pm 0.68$ & $42.31 \pm 0.75$ & $57.56 \pm 0.78$ \\
    RFS \cite{tian2020rethinking} & ResNet-12 & $12.5 \mathrm{M} / 3.5 \times 10^9$ & $73.9 \pm 0.8$ & $86.9 \pm 0.5$ & $44.6 \pm 0.7$ & $\mathbf{6 0 . 9} \pm \mathbf{0 . 6}$ \\
	RE-Net \cite{kang_relational_2021} & ResNet-12 & $12.5 \mathrm{M} / 3.5 \times 10^9$ & $74.51 \pm 0.46$ & $86.60 \pm 0.32$ & - & - \\
	infoPatch \cite{liu_learning_2021} & ResNet-12 & $12.5 \mathrm{M} / 3.5 \times 10^9$ & - & - & $43.8 \pm 0.4$ & $58.0 \pm 0.4$ \\
	MTL \cite{wang2021bridging} & ResNet-12 &  $12.5 \mathrm{M} / 3.5 \times 10^9$ & $69.50 \pm 0.30$ & $84.10 \pm 0.20$ & $42.40 \pm 0.20$ & $57.70 \pm 0.30$ \\	
	SUN \cite{dong2022self} & Visformer-S & $12.4 \mathrm{M} / 1.7 \times 10^8$ & $7 8 . 3 7 \pm 0 . 4 6$ & $8 8 . 8 4 \pm 0 . 3 2$ & - & - \\
	
	Pre-train \cite{chen2023semantic} & Visformer-T & $10.0 \mathrm{M} / 1.3 \times 10^9$ & $71.99 \pm 0.47$ & $85.98 \pm 0.34$ & $43.77 \pm 0.39$ & $59.48 \pm 0.39$ \\
 Meta-AdaM \cite{NEURIPS2023_ce26d216} & ResNet-12 & 12.5M/$3.5 \times 10^9$ & - & - & $41 . 12 \pm 0 . 49$ &  $56 . 14 \pm 0 . 49$ \\
        \hline
        LEP-CLIP \cite{yang2023semantic} &     ResNet-12 & 12.5M/$3.5 \times 10^9$& $80.62\pm0.41$ & $86.22\pm0.33$ & - &  - \\
	
	SP-CLIP \cite{chen2023semantic} & Visformer-T & $10.0 \mathrm{M} / 1.3 \times 10^9$ & $8 2 . 1 8 \pm 0 . 4 0$ & $88.24 \pm 0.32$ & $4 8 . 5 3 \pm 0 . 3 8$ & $\mathbf{6 0 . 12} \pm \mathbf{0 . 4 1}$ \\
	
	\hline
	\textbf{SimpleFSL}	(Ours)  &     Visformer-T & 10.0M/$1.3 \times 10^9$& $\mathbf{84.81} \pm \mathbf{0. 64}$ & $\mathbf{8 8 . 86} \pm \mathbf{0 . 5 5}$ & $\mathbf{48.77} \pm \mathbf{0. 37}$ & $\mathbf{59 . 95} \pm \mathbf{0 . 65}$ \\

	\textbf{SimpleFSL++} (Ours)  &     Visformer-T & 10.0M/$1.3 \times 10^9$& $\mathbf{85.09} \pm \mathbf{0. 64}$ & $\mathbf{8 9 . 10} \pm \mathbf{0 . 31}$ & $\mathbf{49.37} \pm \mathbf{0. 64}$ & $\mathbf{60 . 07} \pm \mathbf{0 . 65}$ \\
	\hline
\end{tabular}
}

\label{tab:2}
\end{table*}

\Cref{tab:1} and \Cref{tab:2} summarize the performances of our proposed two framework SimpleFSL and SimpleFSL++ compared to recent state-of-the-art FSL methods under 5-way 1/5-shot learning tasks on the aforementioned four datasets.
The compared FSL methods include both the single-modal based FSL (visual information only) and multi-modal based FSL (including semantic information of classes). 
Firstly, observed from these experimental results in two tables, our simpleFSL and simpleFSL++ achieve satisfactory performances on all FSL datasets. Especially in the 5-way 1-shot setting, our SimpleFSL and SimpleFSL++ both surpass the SOTA SP-CLIP \cite{chen2023semantic} and LEP-CLIP \cite{yang2023semantic} with substantial accuracy gains. For example, SimpleFSL++ achieve a 4.4\% relative accuracy improvement over the SP-CLIP on miniImageNet, and SimpleFSL achieve a 3.2\% relative accuracy improvement on CIFAR. These obvious performance improvements should be attributed to the explicit utilization of the  pre-trained LM's generalization capacity coordinating with the adaptable learnable prompts, despite the simplicity of both SimpleFSL and SimpleFSL++ frameworks without complex and sophisticated structures. In contrast, previous semantic-based baselines predominantly focus on designing the sophisticated fusion mechanism, ignoring the explicit utilization of LMs.
Secondly, the proposed  SimpleFSL++ consistently outperforms SimpleFSL with better classification accuracy on four datasets, derived from its self-ensemble and self-distillation mechanisms. The detail analysis can be found in Ablation study. 
Thirdly, our evaluation indicates that our SimpleFSL and SimpleFSL++ both have a more pronounced edge in 1-shot than 5-shot classification tasks. This suggests that the visual supervision signal in 5-shot learning is more  abundant  than 1-shot learning , and dominate the model training compared to the semantic supervision signal. These observations align with insights from previous research \cite{xing2019adaptive,chen2023semantic}.

\subsection{Model analysis}

\subsubsection{Ablation study}
\begin{wraptable}[8]{r}{0.5\textwidth}
\vspace{-1.1cm}
	\caption{Ablation study on four datasets under the 5-way 1-shot learning.}
    \centering
\resizebox{0.5\textwidth}{!}{
	\begin{tabular}{l|cccc}
		\hline
		Modules             & Mini & Tired & Cifar & FC100 \\
		\hline 
		Visual backbone     & 65.16   &  72.38    &   71.99    &    43.77  \\
		+ fixed Prompt      &  72.78    & 80.06&   81.12    & 46.46      \\
		+ Learnable Prompt   &   74.80   &  80.21&          84.81  &  48.77 \\
		+ self-ensemble     &   75.14   &  80.30     &          84.92    & 48.93  \\
		+ self-Distillation &  75.59    &  80.52     &   85.09  &  49.37     \\
		\hline
		      
	\end{tabular}
 }

 \label{tab:3}
\end{wraptable}

Fig.\ref{tab:3}  presents the results of the ablation study conducted on four datasets under the 5-way 1-shot learning setting. By combining the visual backbone with the pre-trained LM even with the fixed Prompts, the accuracy can be improved notably. Subsequently, employing the learnable prompts instead of fixed prompts further amplifies performance gains with considerable margins.  Moreover, the incorporation of self-ensemble and self-distillation module in SimpleFSL++ further elevates the accuracy on all four datasets. Collectively, the proposed modules used in SimpleFSL and SimpleFSL++ are instrumental to performance.  The ablation study underscores the significance of explicitly leveraging pre-trained LMs for FSL, which can lead to substantial improvements.

\subsubsection{Prompts analysis}

\begin{wraptable}[9]{r}{0.5\textwidth}
    \centering
    \vspace{-1cm}
\caption{Comparison with different prompt designs on miniImageNet and Cifar under the 5 way 1/5 shot learning.}
\resizebox{0.5\textwidth}{!}{
		\begin{tabular}{lcccc}
   
			\hline
			\multirow{2}{*}{Prompt Type} &      \multicolumn{2}{c}{Mini} & \multicolumn{2}{c}{Cifar } \\ 
			\cline{2-5} & 1-shot &\multicolumn{1}{c}{5-shot} &1-shot & \multicolumn{1}{c}{5-shot} \\ \hline
			
			\hline 
			Dateset-aware   &  $ \mathbf{75.59 }$ & $\mathbf{83.89 }$ & $\mathbf{85 . 09} $ & $\mathbf{89.10 }$  \\
			
			Task-aware  & 73.80 & 83.62 & $84 . 38 $ & $88 . 38 $ \\
			
			Class-aware   & 73.98  & 82.46 & $83.66$ &  88.41 \\
			\hline
		\end{tabular}
}
 \label{tab:4}
\end{wraptable}

Building on the aforementioned discussion, we introduce the learnable prompts instead of previous fiexed prompts in our framework, and ablation study demonstrates its effectiveness for FSL. The learnable prompts we used in SimpleFSL and SimpleFSL++ are all dataset-aware, meaning the learned vector are shared with all classes in a dataset. Inspired by recent progresses in prompt Learning \cite{zhou2022conditional}, we evaluate the class-aware prompts, which are related to specific classes and conditional on the visual prototypes. Further, considering the fact of that few-shot learning consists of many few-shot tasks, we propose task-aware prompts, designed to be unique to each few-shot task and conditional on the visual features of all samples in a FSL task. The more experimental details and settings about the designs of class-aware prompt and task-aware prompt can be found in Appendix. Fig.\ref{tab:4} summarize the 5-way 1/5-shot accuracy comparisons of the three learnable prompt variants. And in Fig.\ref{tab:4}, the fine granularity of modeling decreases from top to bottom sequentially. Unfortunately, the deployments of task-aware or class-aware prompts both do not lead to better performance. These phenomena consist with the conclusion in previous research \cite{zhou2022learning}, which suggests that more fine-grained modeling of prompts might not necessarily enhance downstream task performance. And these potentially are caused by overfitting because the more fine-grained modeling, the more parameters to learning, which is particularly challenging in low-data scenarios. Consequently, we adopt the dataset-aware prompts in our framework and the quest for designing more effective prompts for FSL is a promising avenue for future inquiry.

\subsubsection{Adaptor analysis}
\begin{wraptable}[8]{r}{0.6\textwidth}
\vspace{-0.5cm}
\caption{Comparison with different Adaptors on miniImageNet and Cifar under the 5 way 1/5 shot learning.}
    \centering
\resizebox{0.55\textwidth}{!}{

		\begin{tabular}{lccccc}
			\hline
			\multirow{2}{*}{Adaptor} &     \multirow{2}{*}{Params}& \multicolumn{2}{c}{Mini} & \multicolumn{2}{c}{Cifar } \\ 
			\cline{3-6} && 1-shot &\multicolumn{1}{c}{5-shot} &1-shot & \multicolumn{1}{c}{5-shot} \\ \hline
			
			\hline Linear  & $197 \mathrm{K} $ & $74.82 $ & $83.37 $ & $84.89 $  & $88.96 $ \\
			MLP   & $86 \mathrm{K} $ & $ \mathbf{75.59 }$ & $\mathbf{83.89 }$ & $\mathbf{85 . 09} $ & $\mathbf{89.10 }$ \\
			NLP adapter \cite{houlsby2019parameter}  & $283 \mathrm{K}$ & $74.61 $ & $83.58 $ & $84.62 $ & $89.05 $ \\
			\hline
		\end{tabular}
		}
	
 \label{tab:adaptor}
 \vspace{-0.4cm}
\end{wraptable}

The adaptor in semantic-based few-shot learning plays a significant role \cite{xing2019adaptive,chen2023semantic} which transfers the textual representation into the visual representation space. As illustrated in Fig.\ref{tab:adaptor}, we empirically observe that different adaptors bear some influence on the classification performance. The linear adaptor is frequently used in previous works \cite{xing2019adaptive,chen2023semantic}. The 2-layer MLP with a bottleneck structure are used in our framework, with a small number of hidden dimension. Inspired by the NLP adaptors \cite{houlsby2019parameter}, we also evaluate the combination of the bottleneck structure and a linear layer with the residual connection \cite{he2016deep}. The experimental results demonstrate that our MLP w/ bottleneck has slight advantage, which may own to its less parameters. The experimental details about the three adaptors can be found in the Appendix.

\subsubsection{Fusion mechanism}
\begin{wraptable}[10]{r}{0.5\textwidth}
\vspace{-1.1cm}
\caption{Comparison with different fusion mechanisms on miniImageNet and Cifar under the 5 way 1/5 shot learning.}
	\centering
 \resizebox{0.48\textwidth}{!}{
		\begin{tabular}{lcccc}
			\hline
			\multirow{2}{*}{Fusion} &      \multicolumn{2}{c}{Mini} & \multicolumn{2}{c}{Cifar } \\ 
			\cline{2-5} & 1-shot &\multicolumn{1}{c}{5-shot} &1-shot & \multicolumn{1}{c}{5-shot} \\ \hline
			
			\hline 
                SP-CLIP \cite{chen2023semantic} & $72.41$ & $83.23$ & $8 2 . 1 8 $ & $88.24 $ \\
			Add   &  $ 75.59 $ & $\mathbf{83.89 }$ & $\mathbf{85 . 09} $ & $\mathbf{89.10 }$  \\
			Concat  &  $74.42 $ & $83.00 $ & $83.96 $ & 88.80 \\
			Attention \cite{xing2019adaptive}  & $\mathbf{75.61} $ & $83.75 $ & $84.39 $ & $\mathbf{89.10} $ \\
			\hline
		\end{tabular}
            }
	
 \label{tab:fusion}
\end{wraptable}

In both SimpleFSL and SimpleFSL++, we adopt a straightforward addition operation to fuse the visual representation and semantic representation for multi-modal feature fusion. We also re-implement Simple++ with different fusion mechanism, including the Concatenation and Attention \cite{xing2019adaptive}, and experimental results are summarized in \Cref{tab:fusion}. Here we use the SP-CLIP \cite{chen2023semantic} as the baseline , which design a complex fusion mechanisms to insert semantic representations into the feature extractor. The results in Fig.\ref{tab:fusion} reveal negligible differences in performance between employing Addition or Attention. 
This suggests that the simple fusion operation is benefit for the utilization of pre-trained LMs, and the obtained semantic feature can directly assist to classify without passing the complex architecture which may hurt the generalization capability of semantic features. 
For simplicity, we adopt the add operation in our framework. The experimental details about the Concatenation and Attention are described in the Appendix.


\subsubsection{Hyper-parameters analysis}





\begin{figure} [t]
	\centering
	\subfloat[1-shot on mini]{
		\includegraphics[scale=0.3]{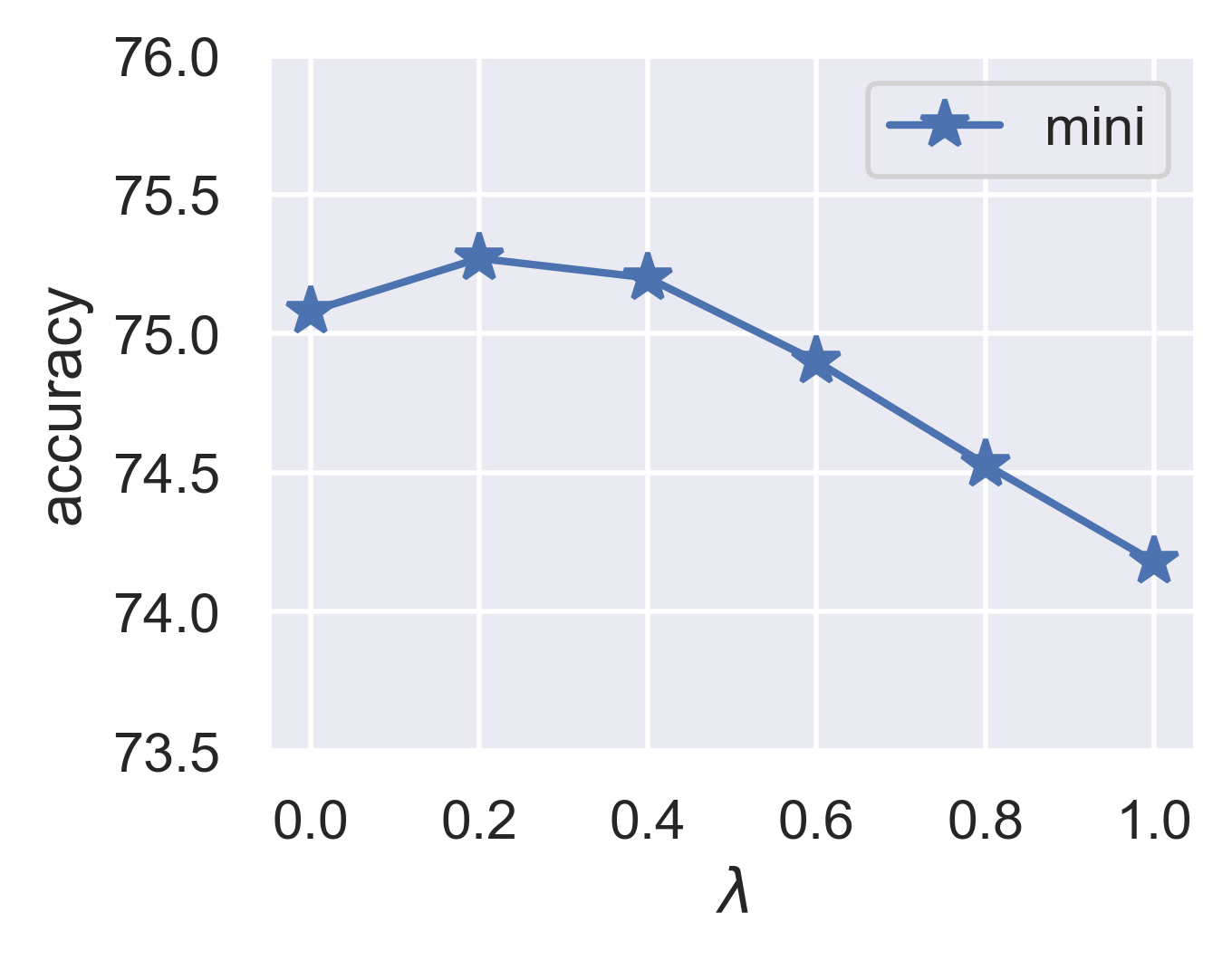}}
	\subfloat[1-shot on CIFAR]{
		\includegraphics[scale=0.3]{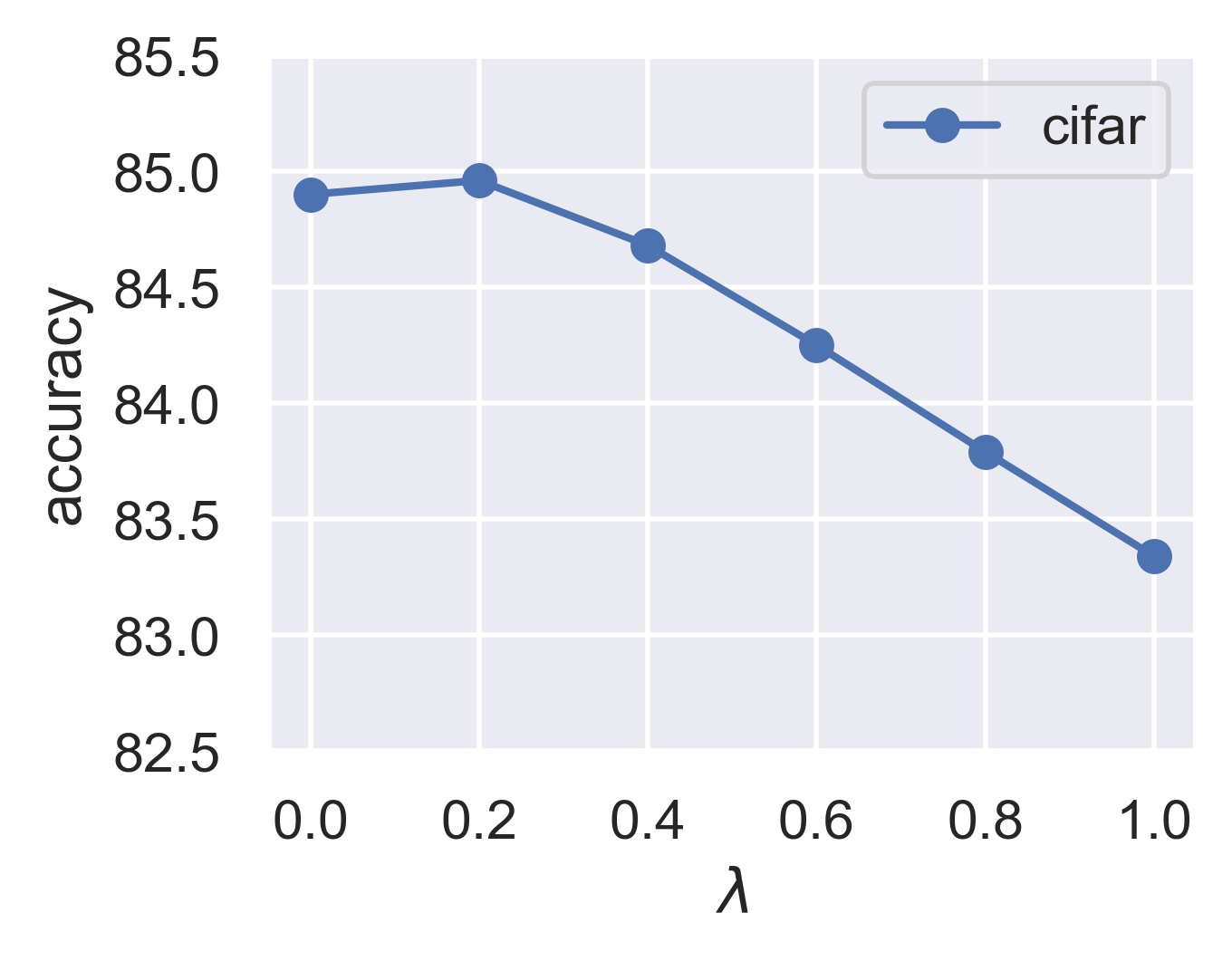}}
	\subfloat[5-shot on mini]{
		\includegraphics[scale=0.3]{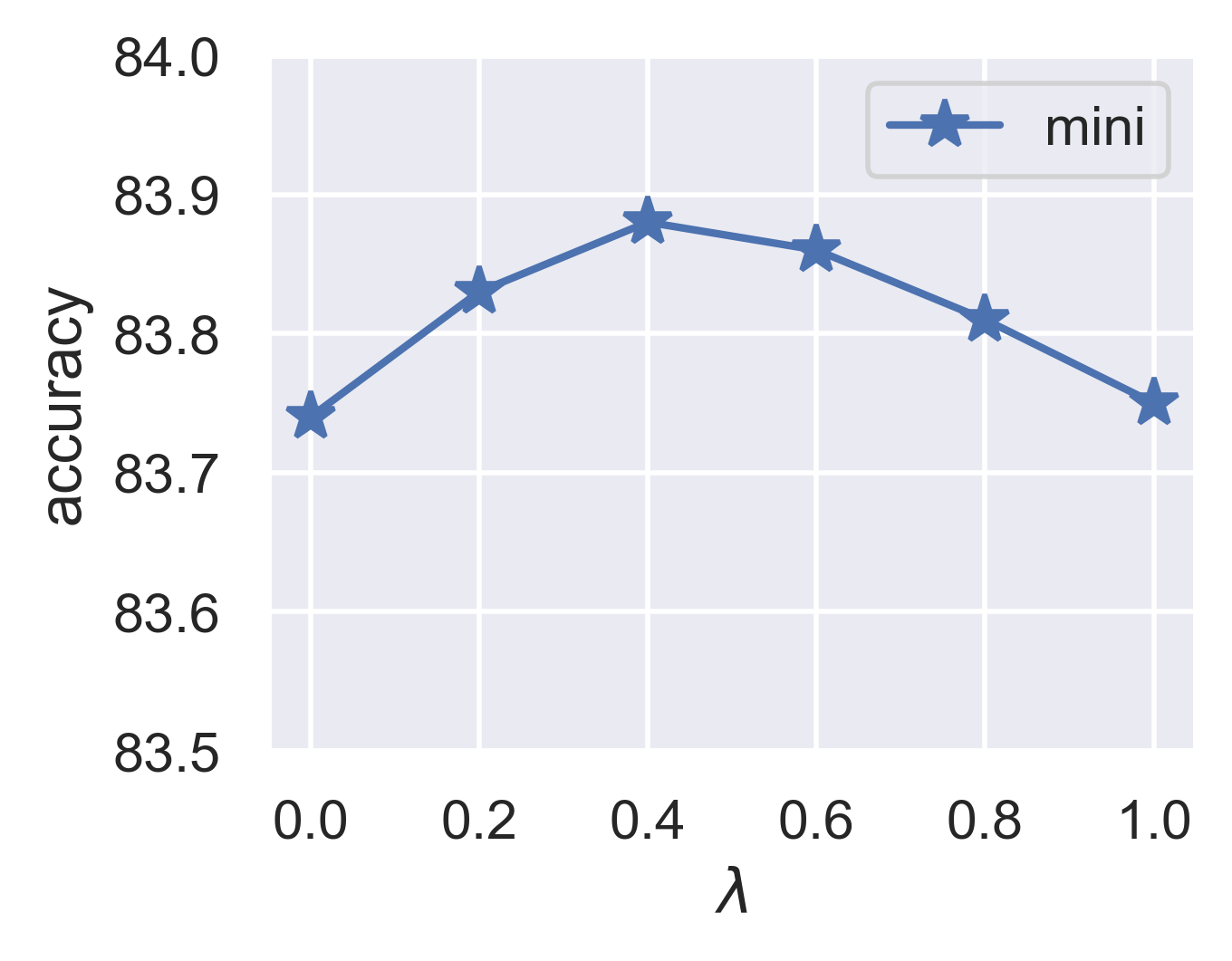} }
	\subfloat[5-shot on CIFAR]{
		\includegraphics[scale=0.3]{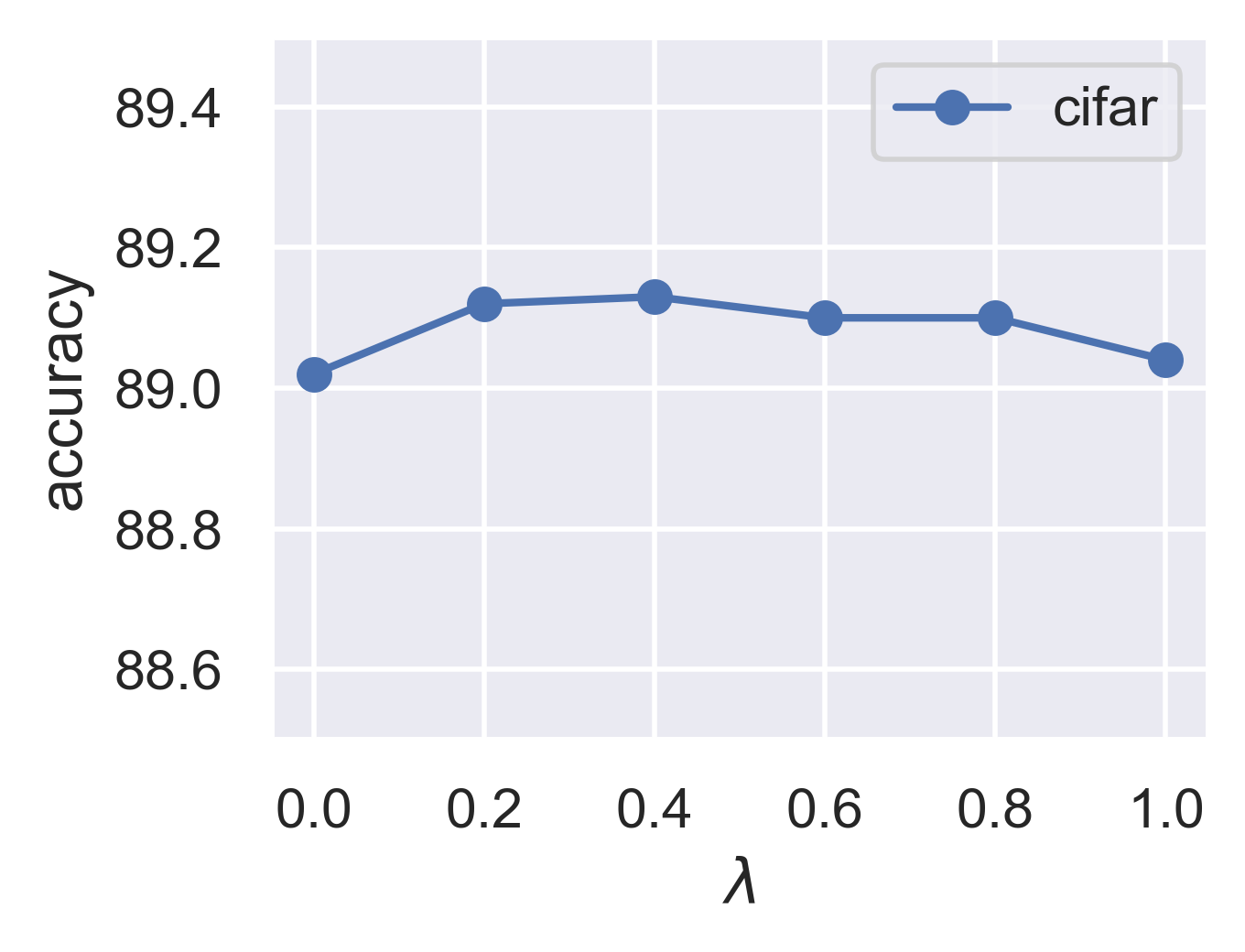}}

	\caption{Evaluation of  different weighting factor $\lambda$ on miniImageNet and CIFAR-FS.}
	\label{fig:3} 
 \vspace{-0.4cm}
\end{figure}

\begin{figure} [t]
	\centering
	\subfloat[1-shot on mini]{
		\includegraphics[scale=0.3]{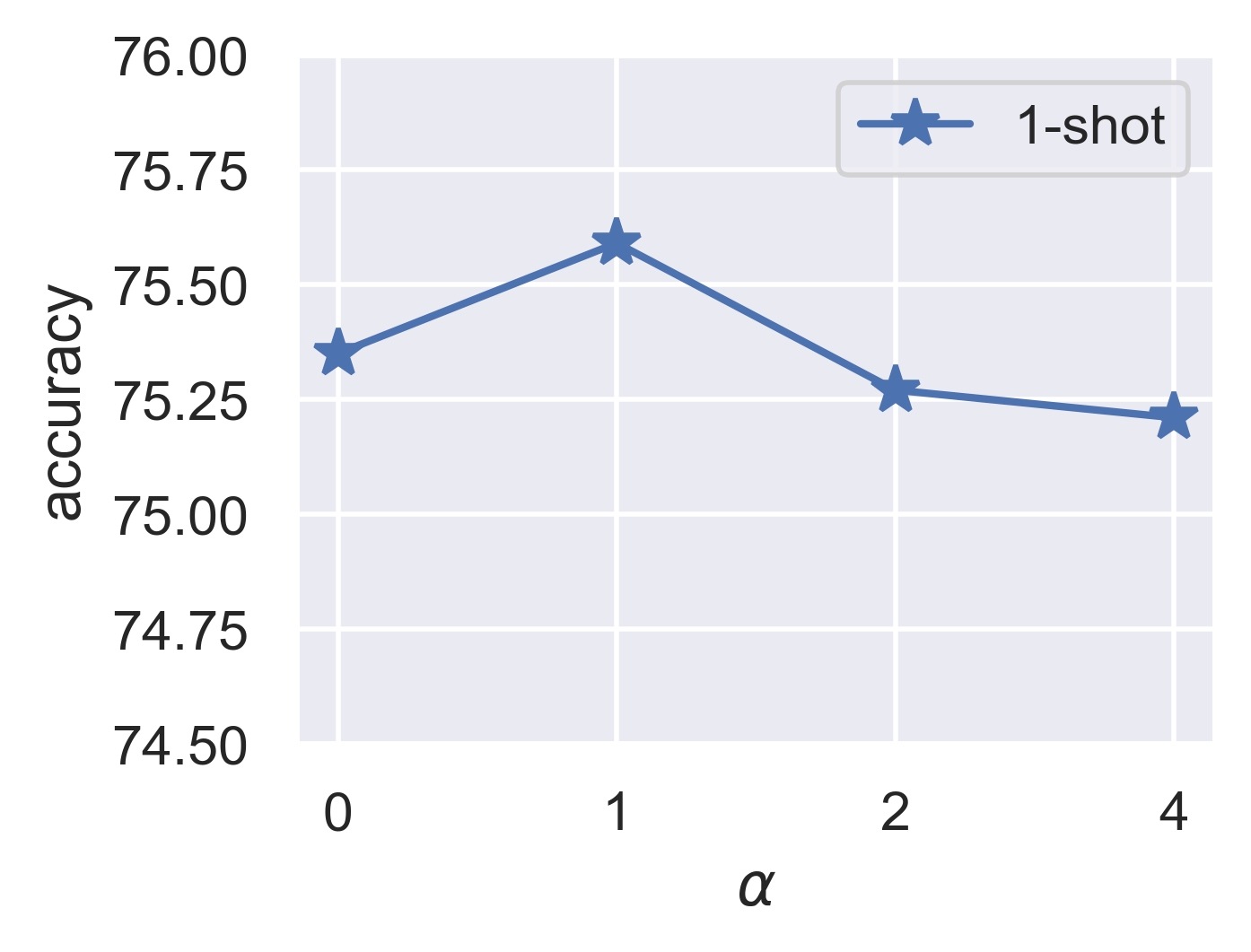}}
	\subfloat[1-shot on CIFAR]{
		\includegraphics[scale=0.3]{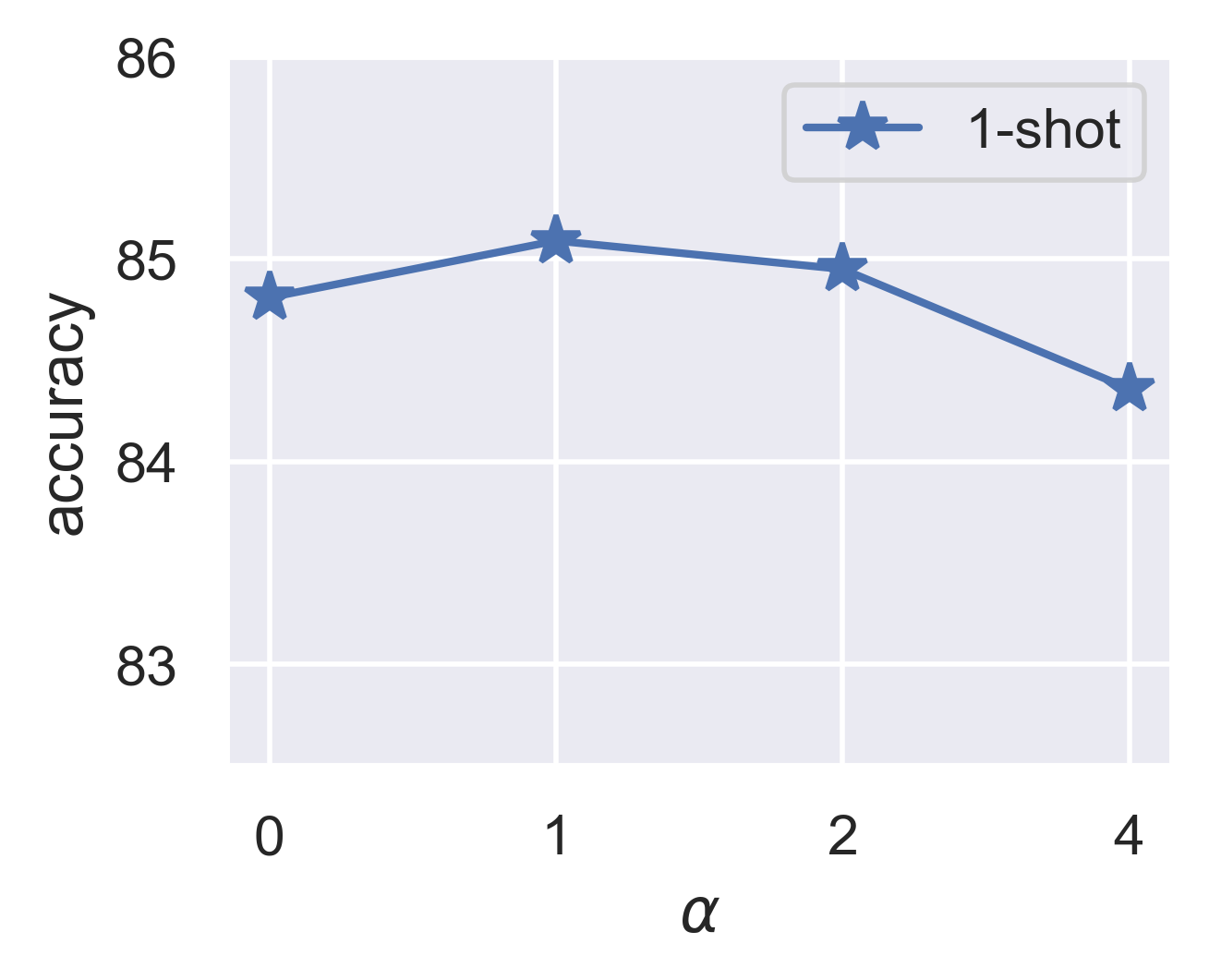}}
	\subfloat[5-shot on mini]{
		\includegraphics[scale=0.3]{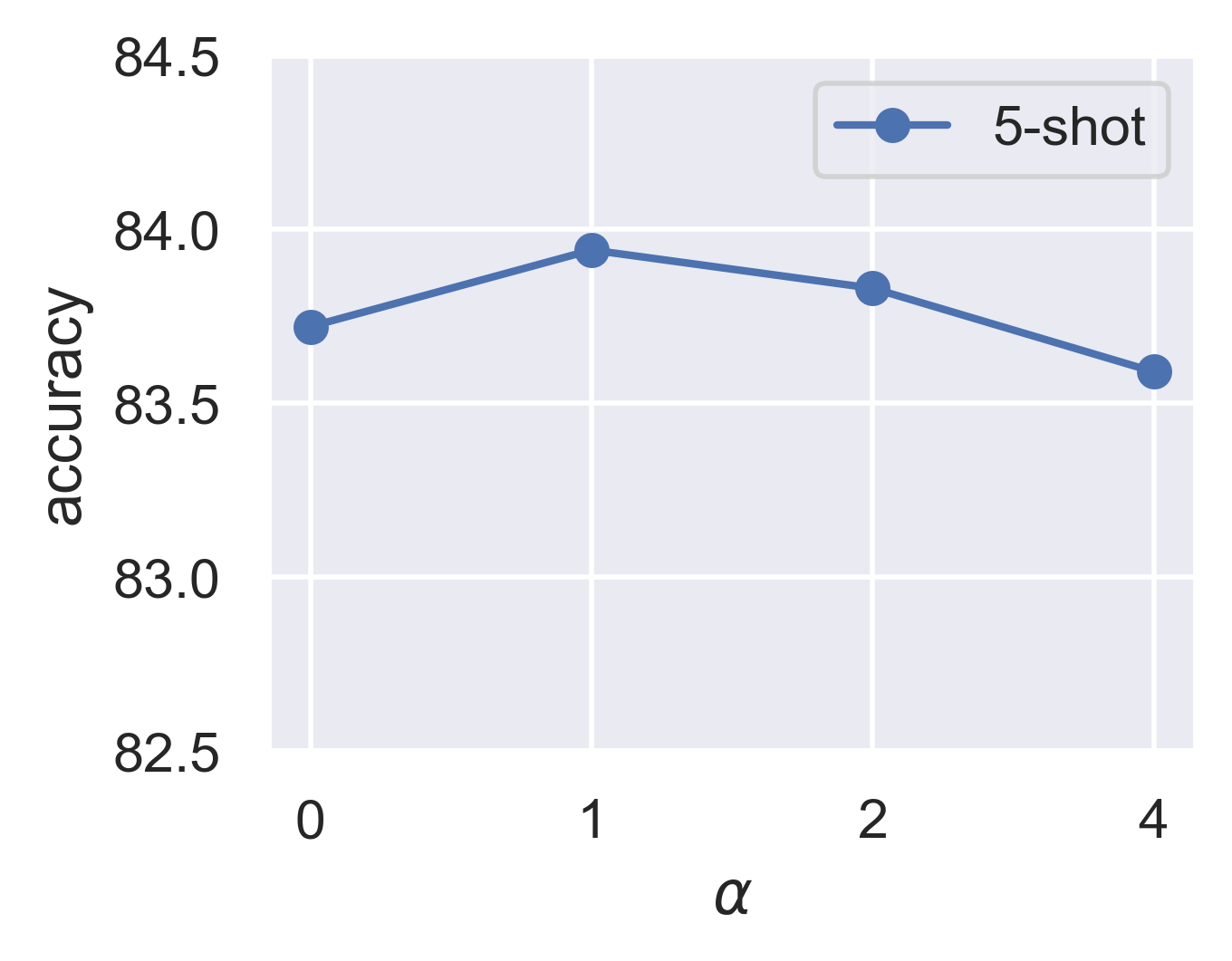} }
	\subfloat[5-shot on CIFAR]{
		\includegraphics[scale=0.3]{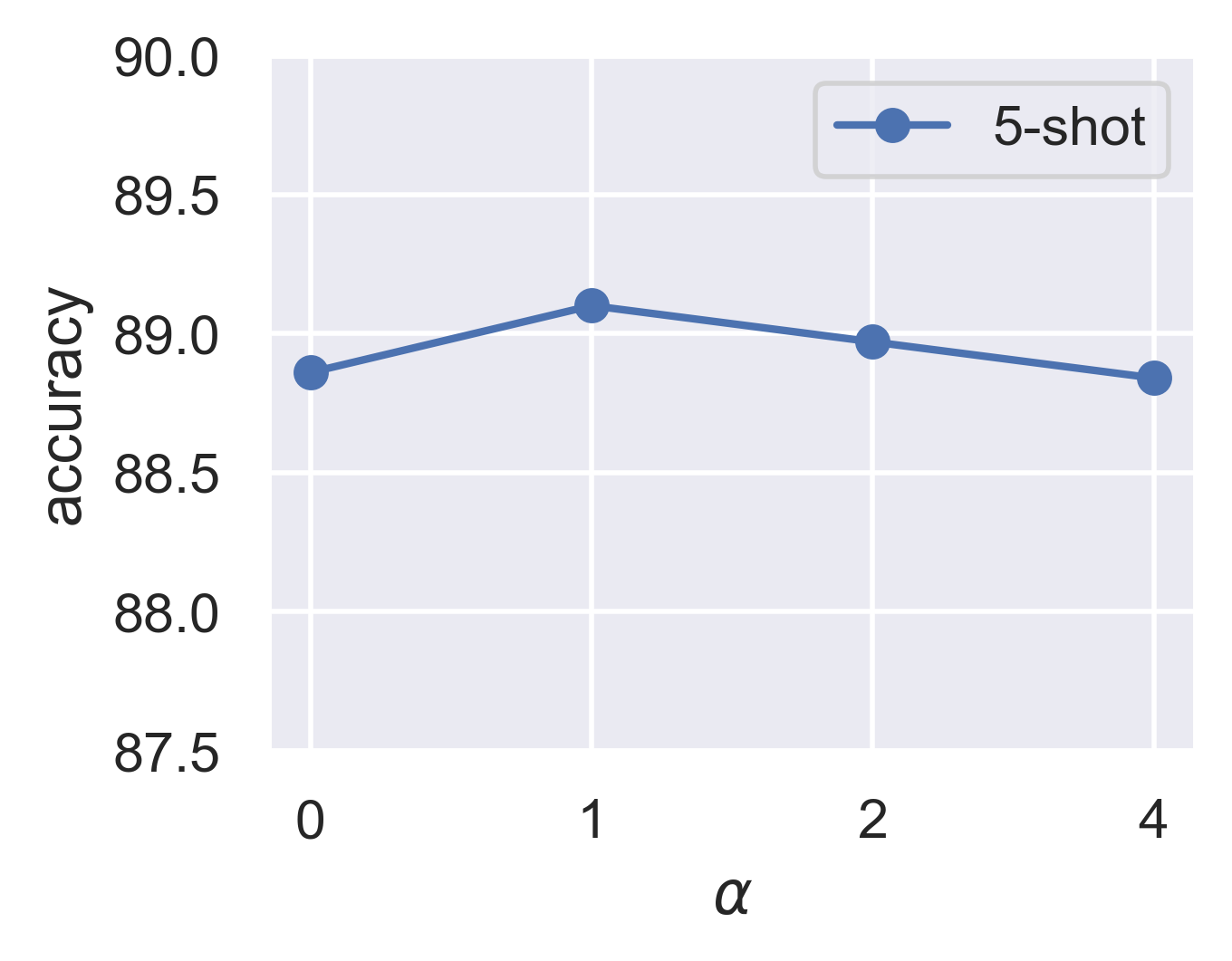}}

	\caption{Evaluation of  different Distillation weight $\alpha$ on miniImageNet and CIFAR-FS.}
	\label{fig:4} 
\end{figure}

The ablation study highlights that the self-ensemble and self-distillation module both confer an additional performance enhancement. Then, we delve deeper into the impact of these modules on the performance in SimpleFSL++.  As delineated previously, $\lambda$ servers as the weighting factor to control the weighting of two classifiers in Eq.\ref{eq:11}, and $\alpha$ denotes the weighting factor of the distillation loss in Eq.\ref{eq:13}. 
Fig.\ref{fig:3} and Fig.\ref{fig:4} summarize the performance with varied  $\lambda$ and $\alpha$ under 5-way 1/5-shot learning on the miniImageNet and CIFAR-FS, respectively. We simultaneously observe that the utilization of self-ensemble and distillation module with an appropriate weight ameliorates performance on these datasets. While an excessively high weighting value may not yield further benefits. Specially, an observation of Fig.\ref{fig:3} reveals that the ideal $\lambda$ setting for 1-shot learning is lower compared to that in 5-shot learning.  This discrepancy can potentially be also attributed to the fact that 5-shot learning offers a more stonger visual supervision signal in comparison to 1-shot learning, thereby diminishing the relative importance of semantic guidance, as discussed in Sec \ref{sec:52}.





\begin{wrapfigure}[20]{r}{0.48\textwidth}
    \centering
    \vspace{-17pt}
        \subfloat[backbone on mini and CIFAR]{
		\includegraphics[scale=0.18]{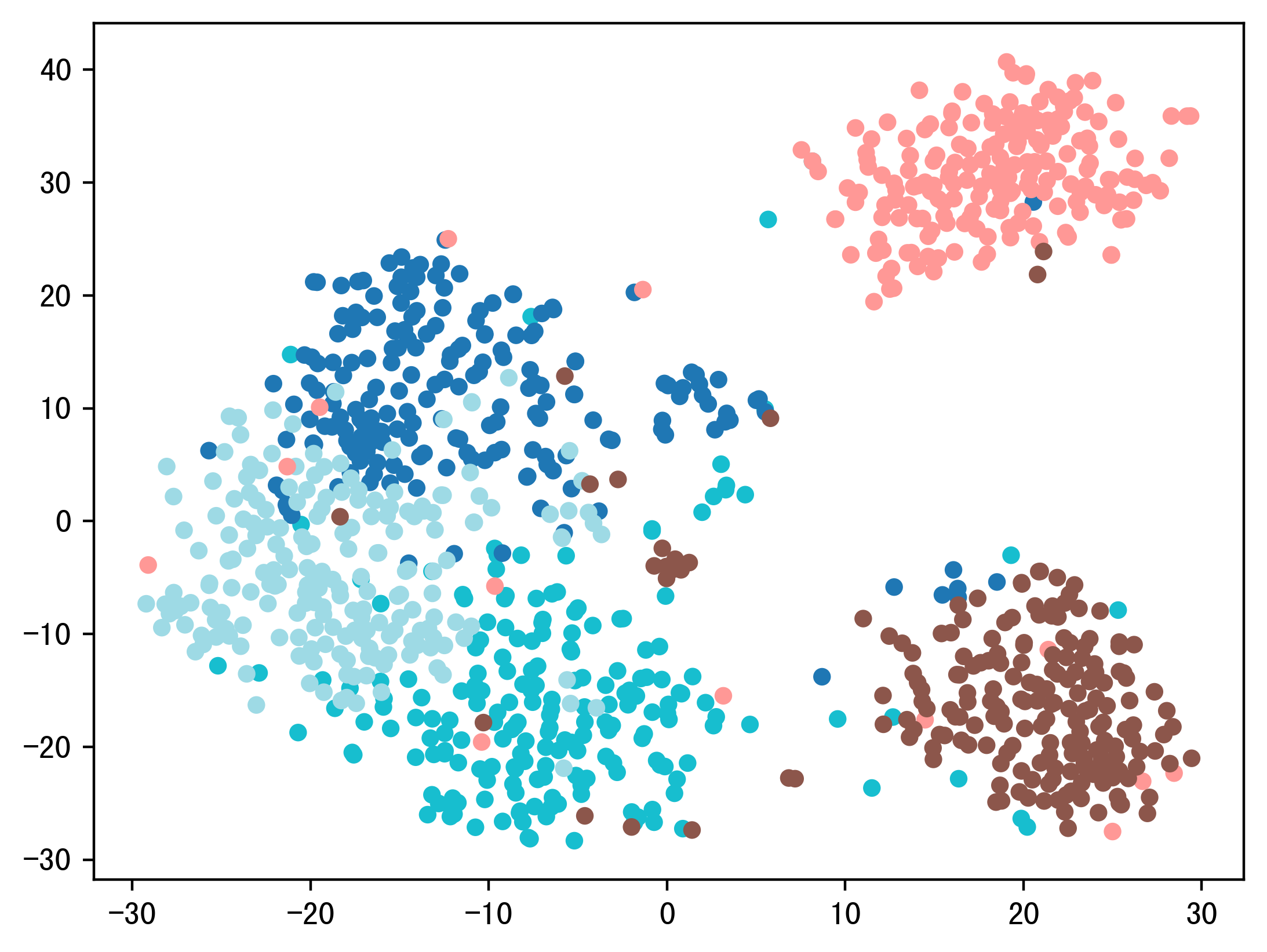}
		\includegraphics[scale=0.18]{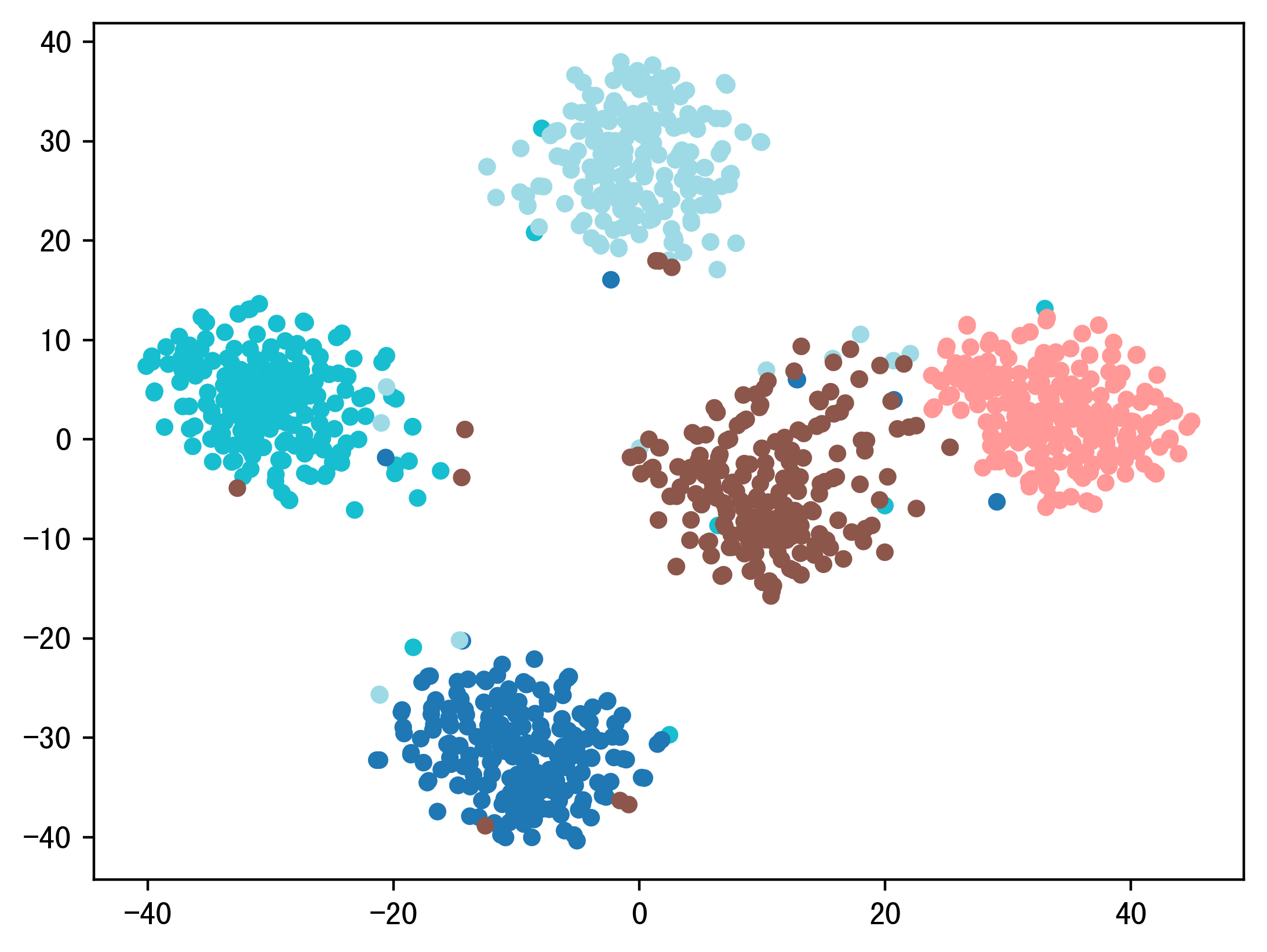}}\\
  
        \subfloat[SP-CLIP on mini and CIFAR]{
		\includegraphics[scale=0.18]{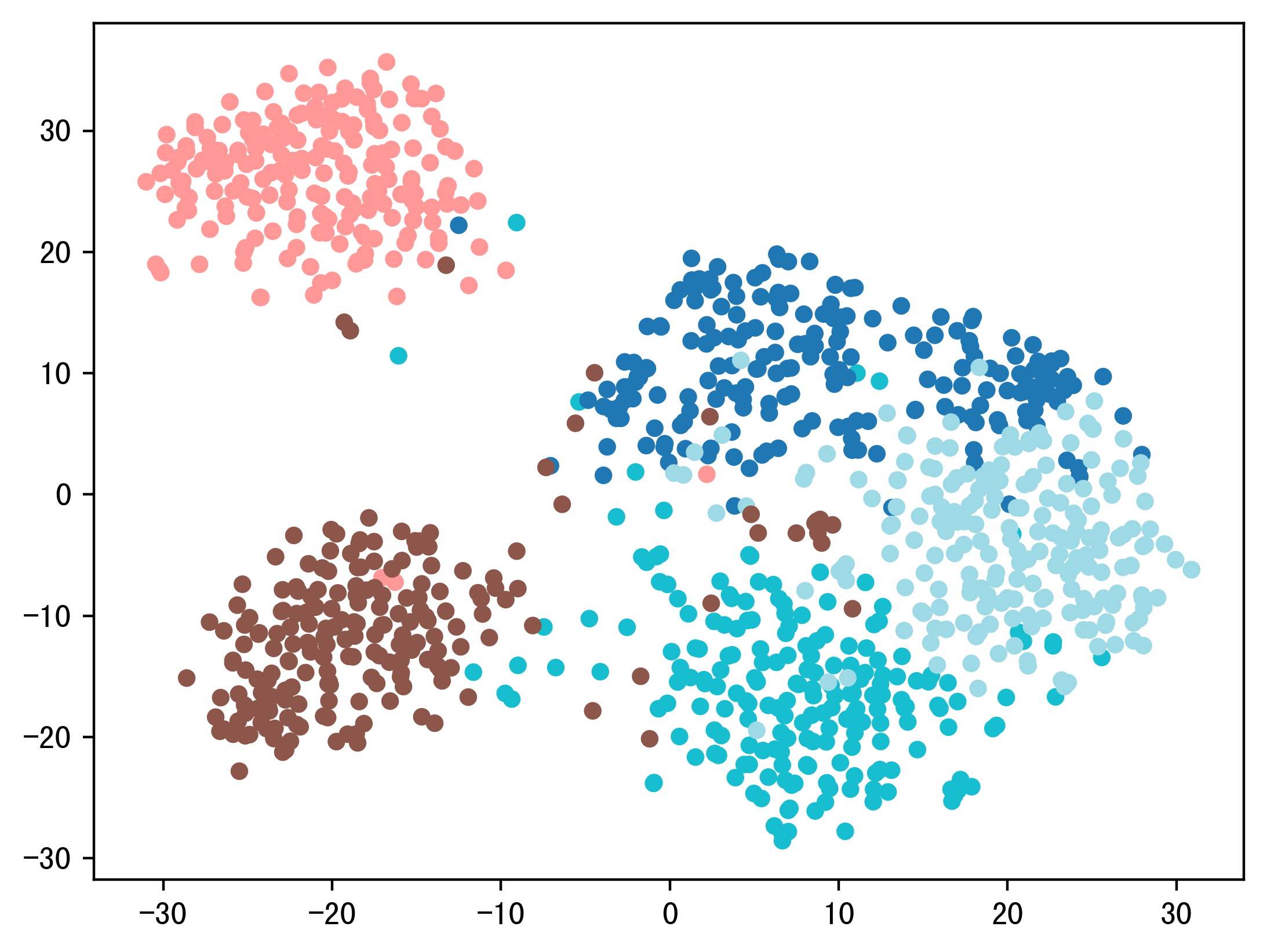}
		\includegraphics[scale=0.18]{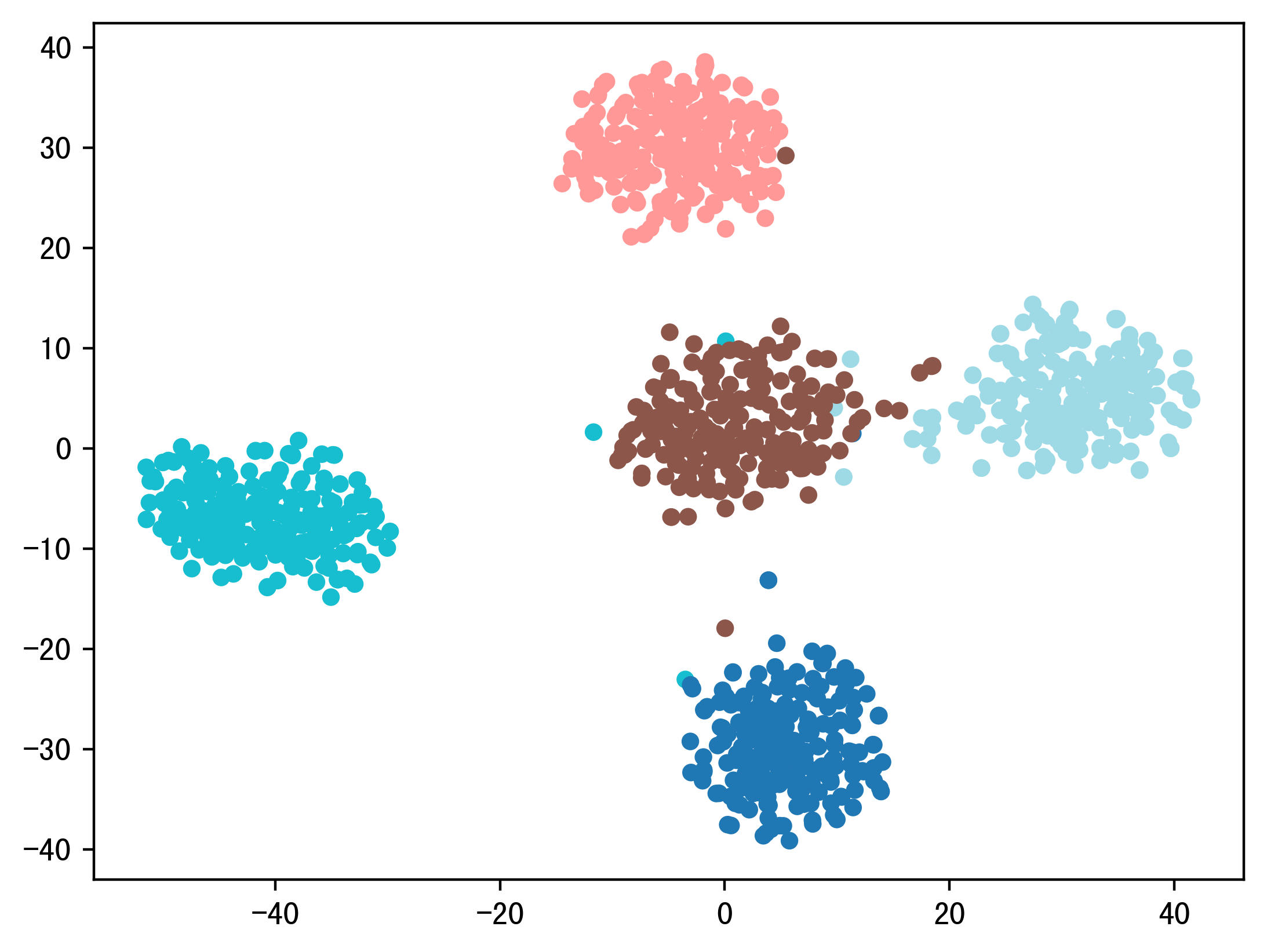}}\\
   
	\subfloat[Ours on mini and CIFAR]{
		\includegraphics[scale=0.18]{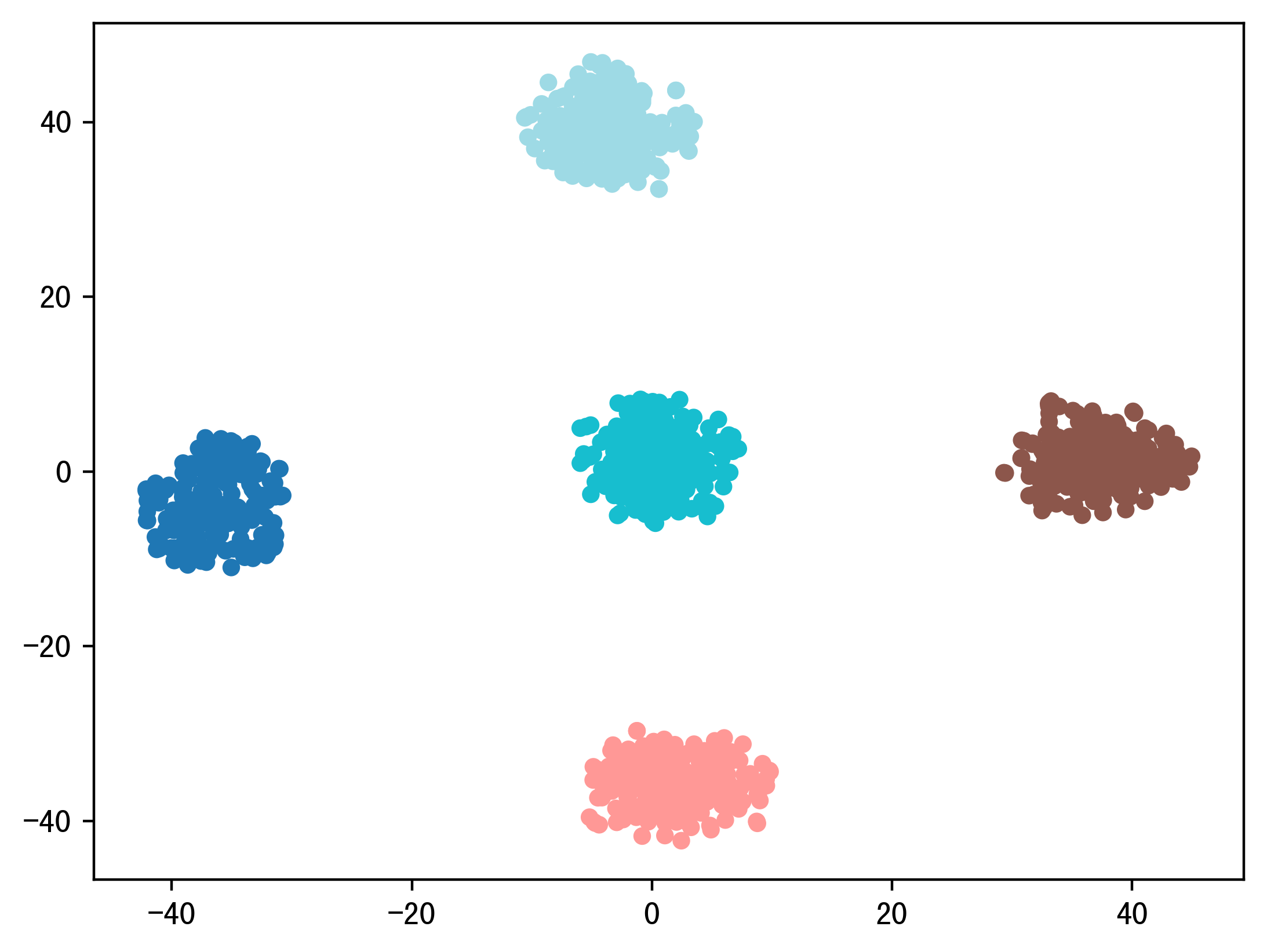} 
		\includegraphics[scale=0.18]{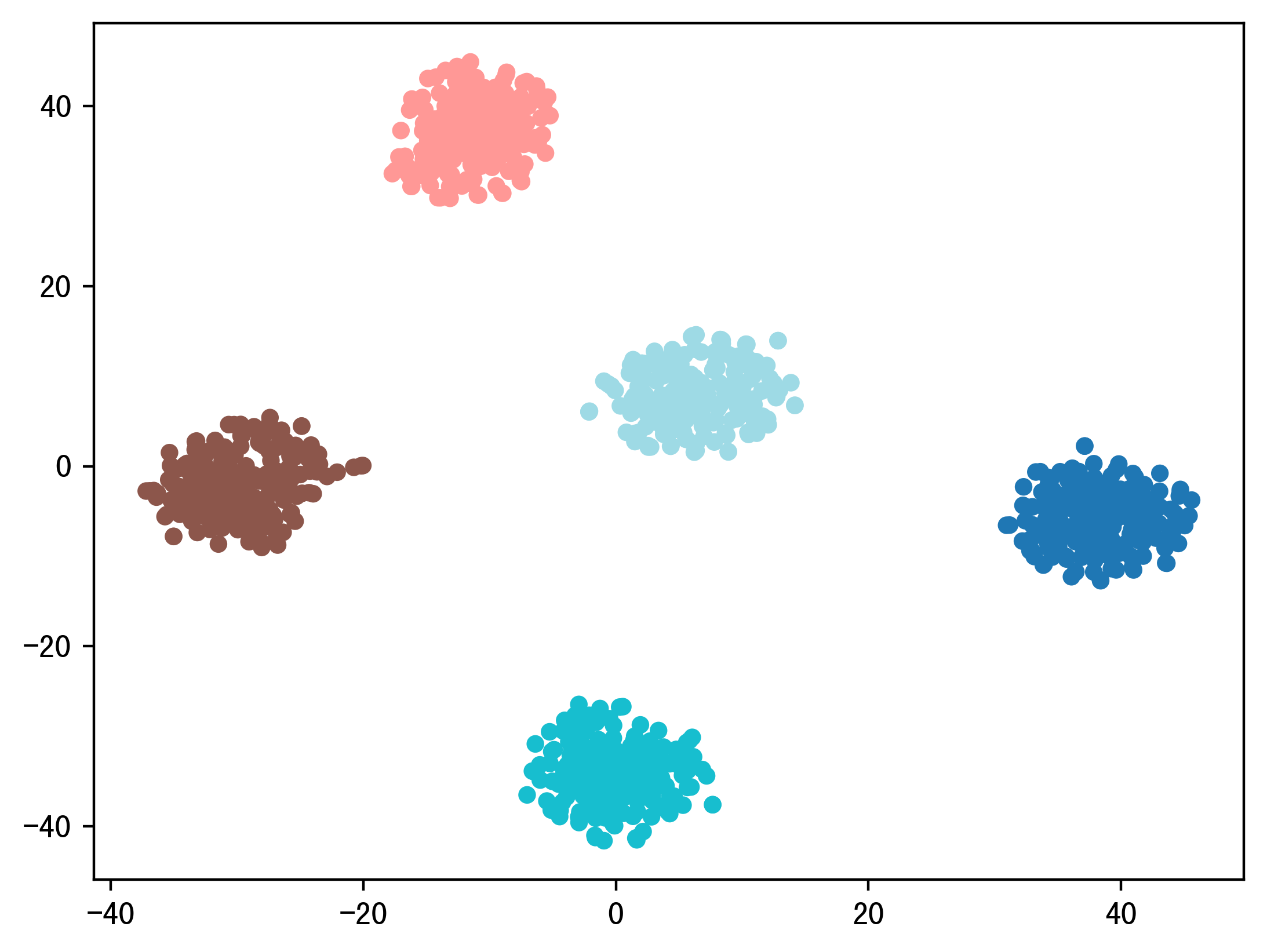} }
    \caption{t-SNE visualizations of the representation of different methods.}
    \label{fig:5}
\end{wrapfigure}

\vspace{-0.3cm}
\subsubsection{Visualization}
In Fig.\ref{fig:5}, we present the t-SNE \cite{van2008visualizing} visualizations of the last-layer representation prior to classifier-3 in SimpleFSL++, compared to a sole visual backbone without semantic guidance and SP-CLIP \cite{chen2023semantic}. We visualize the support set in a randomly sampled 5-way 200-shot task from $\mathbf{D_{novel}}$ in miniImageNet and CIFAR-FSL, with distinct classes denoted by different colors.   
All plots in Fig.\ref{fig:5} collectively demonstrate that ours SimpleFSL++ achieves more distinct class separability, notwithstanding its straightforward architecture. In comparison, the more intricate fusion mechanism of SP-CLIP does not significantly surpass the discriminability provided by a visual backbone without semantic guidance.

\vspace{-0.4cm}
\section{Conclusion}
\vspace{-0.2cm}
In this paper, we focus on the few-shot image classification task, and emphasize the generalization capability of the pre-trained language model in the few-shot learning, which is usually underestimated in previous works. To harness this capability, we propose a straightforward and efficacious framework for few-shot learning tasks, instead of designing the intricate and complex architectures. And we directly add the visual feature with the textual feature with adaptable learnable prompts, allowing for flexible accommodation to various datasets. Further, we apply the self-ensemble and self-Distillation to bring the additional performance boost.  Our extensive experiments conducted across four few-shot datasets demonstrate that our proposed framework consistently delivers promising  results, with particularly notable performance in the 1-shot learning task. The exploration of prompt design, tailored to optimize few-shot learning deserves further investigation in the future. 



\clearpage  

%
%
\bibliographystyle{splncs04}
\bibliography{main}

\clearpage 

\appendix
\renewcommand\thefigure{S\arabic{figure}}
\renewcommand\thetable{S\arabic{table}}

\setcounter{figure}{0}
\setcounter{table}{0}
\setcounter{equation}{0}

\begin{center} \Large
\textbf{Supplementary Material }   
\end{center}

\section{Ablation Study of pre-trained LM}
In this section, we aim to explore the impact  that various pre-trained Language Models (LMs) exert on our framework's performance, as have done in previous Semantic-based few-shot learning works \cite{xing2019adaptive,liu2021cross,yang2023semantic,chen2023semantic}. 
As delineated in the methodology section, we utilize pre-trained LMs as the semantic feature extractors, and the compared LMs include: GloVe \cite{pennington2014glove}, ALBERT \cite{lan2019albert}, RoBERTa \cite{liu2019roberta} and CLIP \cite{radford2021learning}.
For a fair comparison, we deploy these LMs with fixed prompts (e.g., a photo of a cat) here in our proposed SimpleFSL framework.
As reported in Tab.\ref{table:S1}, we first observe that our proposed framework consistently bring the significantly performance improvement with various LMs, compared to the purely visual backbone without using LMs.  This outcome demonstrably affirms the efficacy of our framework. Moreover, we observe that incorporating CLIP leads to superior performance in most scenarios compared to the other LMs. This observation is in harmony with the findings of recent studies \cite{yang2023semantic,chen2023semantic}, which may be attributable to that CLIP’s features particularly well-suited for aligning semantic and visual representations.
Therefore, we opt CLIP as our default semantic feature extractor in our implementations.

\begin{table}[ht] 
	\centering
 \caption{Performance comparison with different language models in  SimpleFSL on miniImageNet and CIFAR-FS.}
	\setlength{\tabcolsep}{2.5mm}{
		\begin{tabular}{ccccc} 
			\hline
			\multirow{2}{*}{Language Model} &      \multicolumn{2}{c}{miniImageNet} & \multicolumn{2}{c}{CIFAR-FS} \\ 
			\cline{2-5} & 1-shot &\multicolumn{1}{c}{5-shot} &1-shot & \multicolumn{1}{c}{5-shot} \\ \hline
            w.o. LM     & 65.16   &  81.22&   71.99    &    85.98\\
                 GloVe \cite{pennington2014glove}  & $67.89 $ & $81.73 $ & $ 80.37$  & $ 88.57$ \\
			ALBERT \cite{lan2019albert} & $67.31 $ & $81.50 $ & $80.20 $  & $88.08 $ \\
			RoBERTa \cite{liu2019roberta} & $ 70.11$ & $82.26$ & $ \mathbf{81.26}$ & $88.72 $ \\
			CLIP  \cite{radford2021learning} & $\mathbf{72.78} $ & $ \mathbf{83.34} $ & $81.12 $ & $\mathbf{88.86} $ \\
			\hline
		\end{tabular}
		}
            \vspace{0.3cm}
\label{table:S1}
	
\end{table}

\section{Additional Visualization}
In the section of Visualization, we have presented the t-SNE visualizations  of representations of different methods, sampled from the miniImageNet and CIFAR-FS datasets. And we provide more visualization examples from the tieredImageNet and FC100 datasets. Similarly, we visualize the support set in a randomly sampled 5-way 200-shot task from $\mathbf{D_{novel}}$ in tieredImageNet and FC100, with each class represented by a unique color. Notably, we observe that ours SimpleFSL++ continues to yield impressive performance in these two datasets, presenting more distinct features  in comparison to  other baselines. These visualizations reinforce our earlier observation that the more intricate fusion mechanism utilized by SP-CLIP \cite{chen2023semantic} fails to offer a significant advantage in discriminability over a visual backbone without semantic guidance.


%

\begin{figure}[ht]  
\centering
    \subfloat[backbone on tieredImageNet and FC100]{
		\includegraphics[scale=0.3]{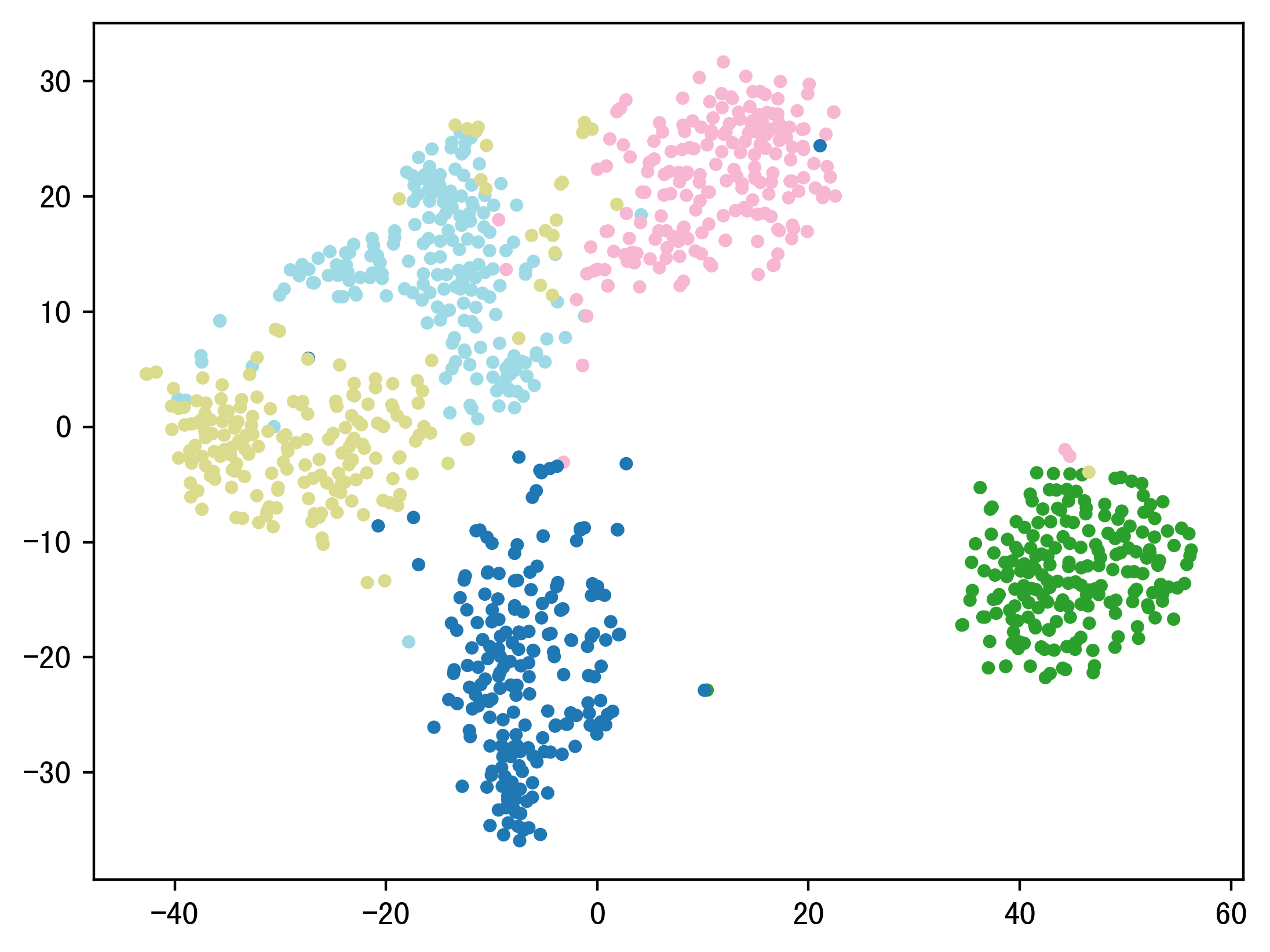}
		\includegraphics[scale=0.3]{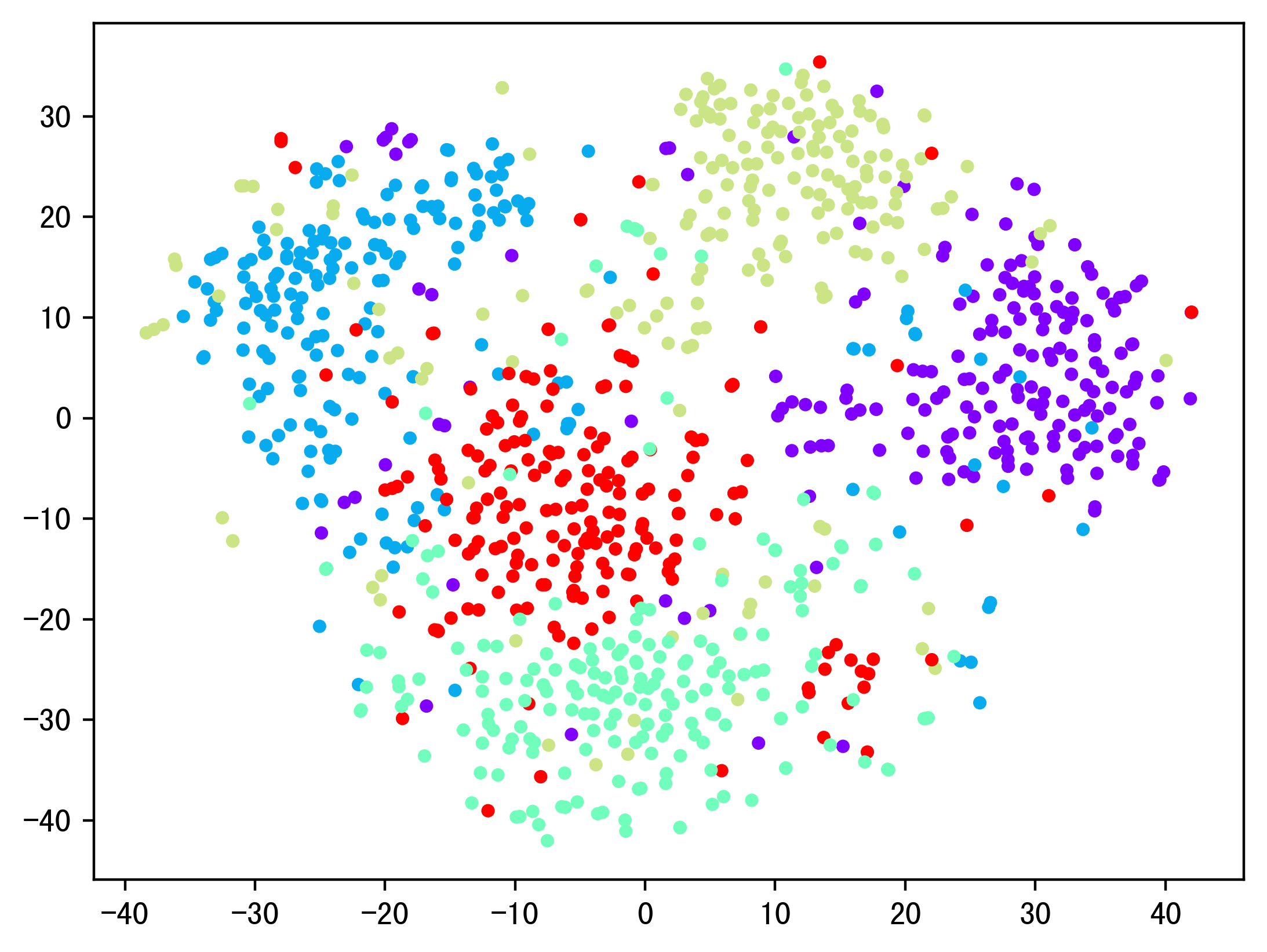}}\\
  
        \subfloat[SP-CLIP on tieredImageNet and FC100]{
		\includegraphics[scale=0.3]{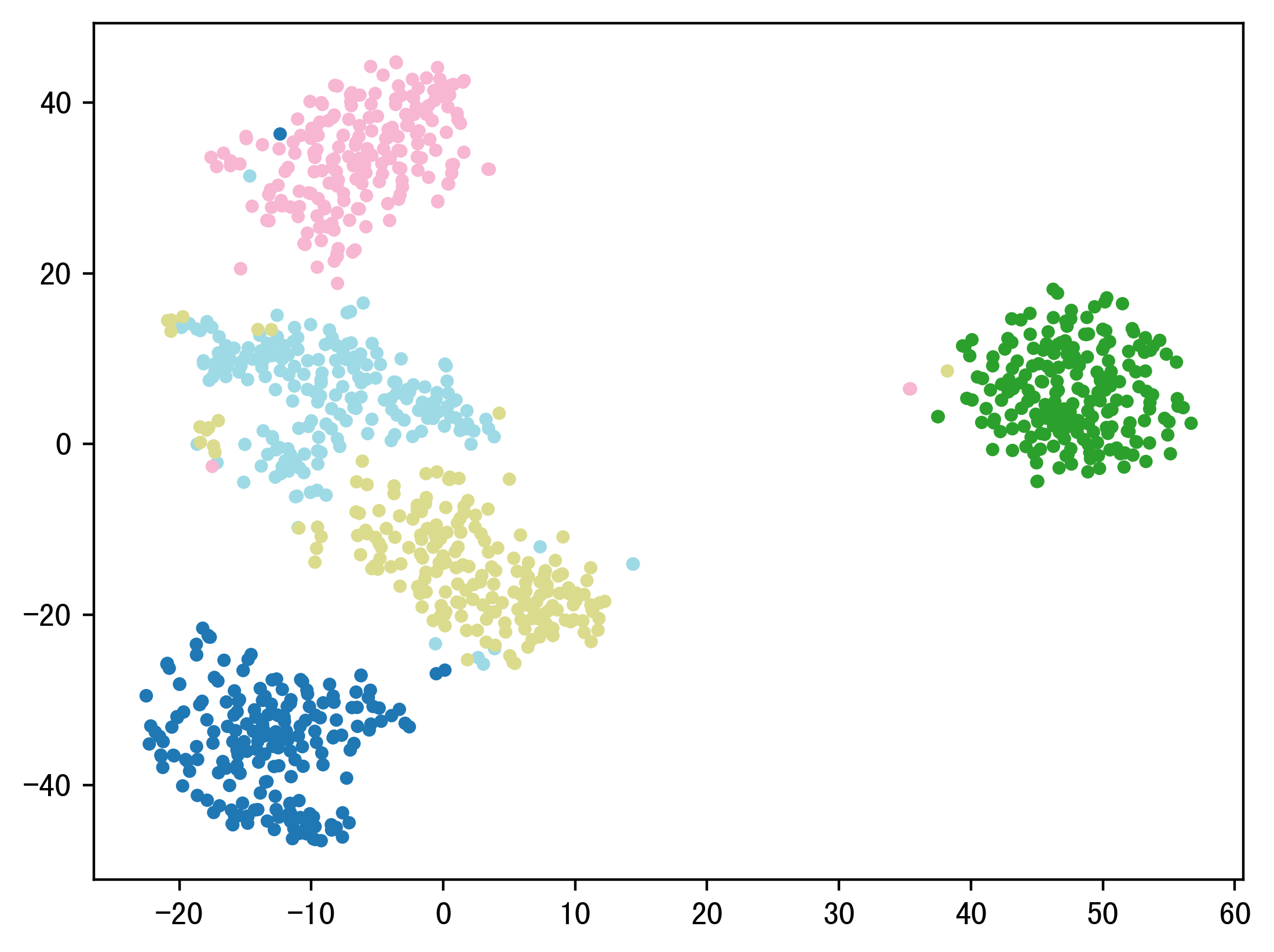}
		\includegraphics[scale=0.3]{fig/FC100_2tsne_vis.png}}\\
   
	\subfloat[Ours on tieredImageNet and FC100]{
		\includegraphics[scale=0.3]{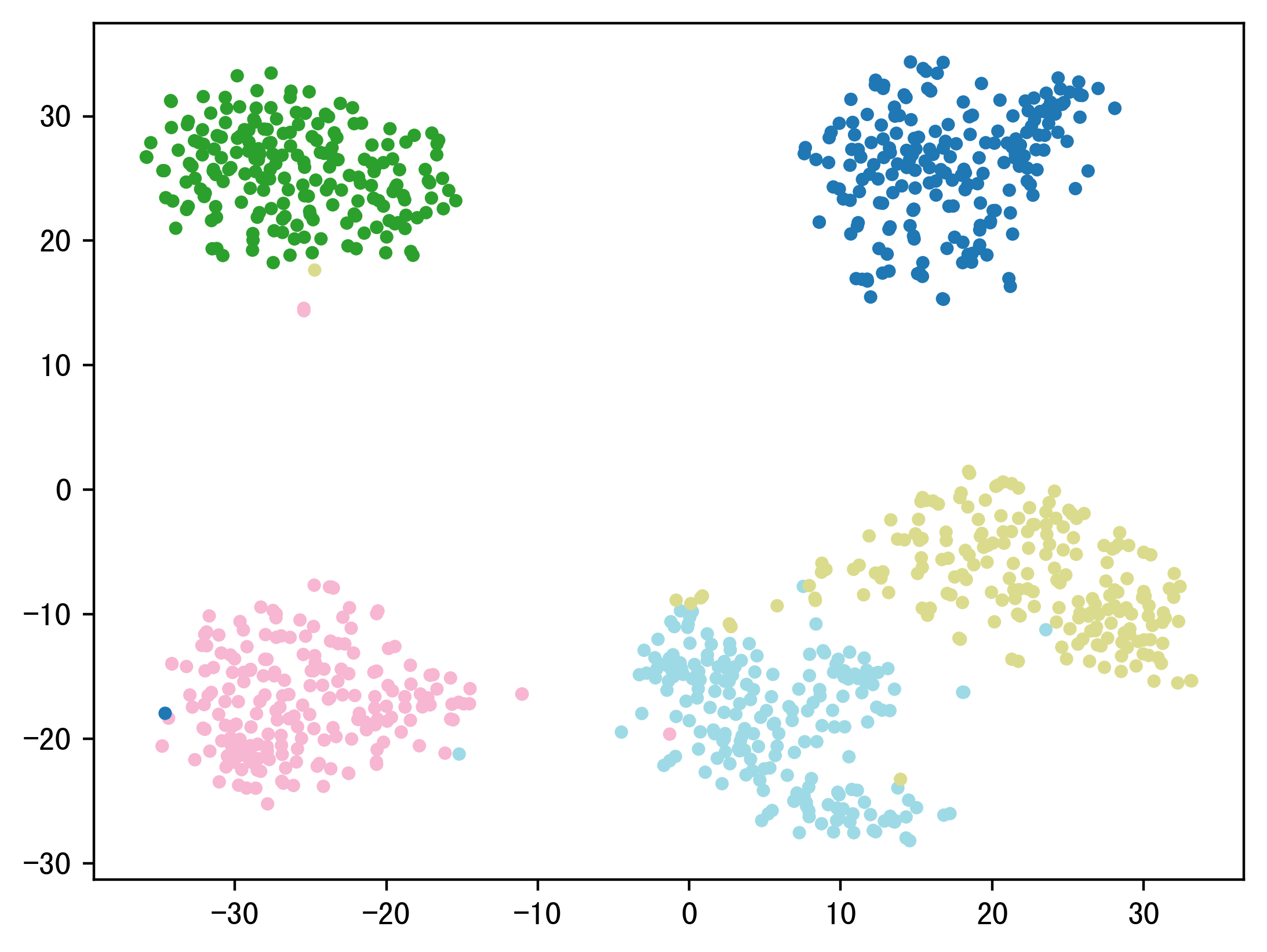} 
		\includegraphics[scale=0.3]{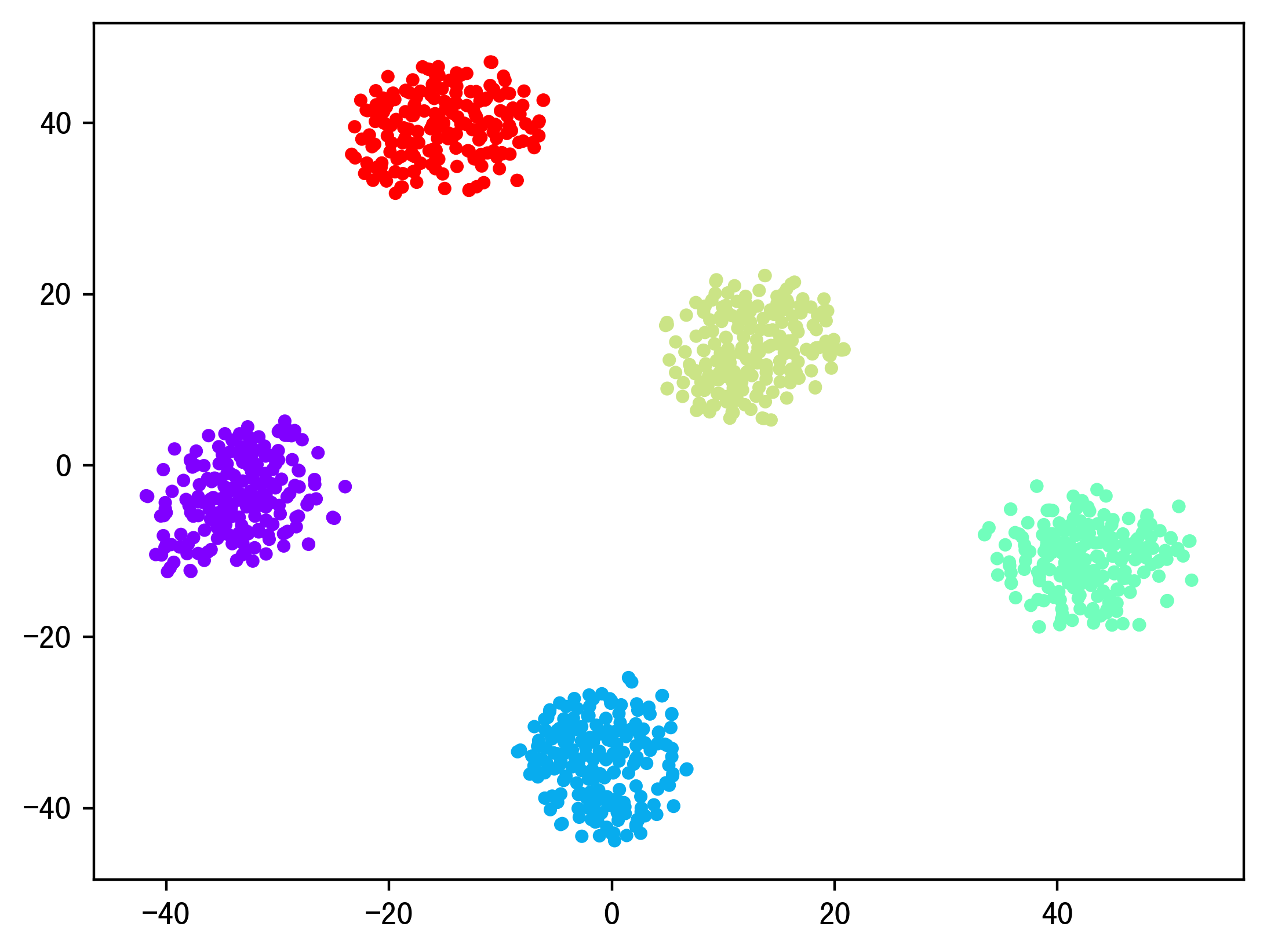} }
    \caption{t-SNE visualizations of the representation of different methods.}
    \label{fig:S1}
\end{figure}

\section{Few-Shot Baselines with LM}
\begin{figure*}[h]
	\centering
	\includegraphics[width=13cm]{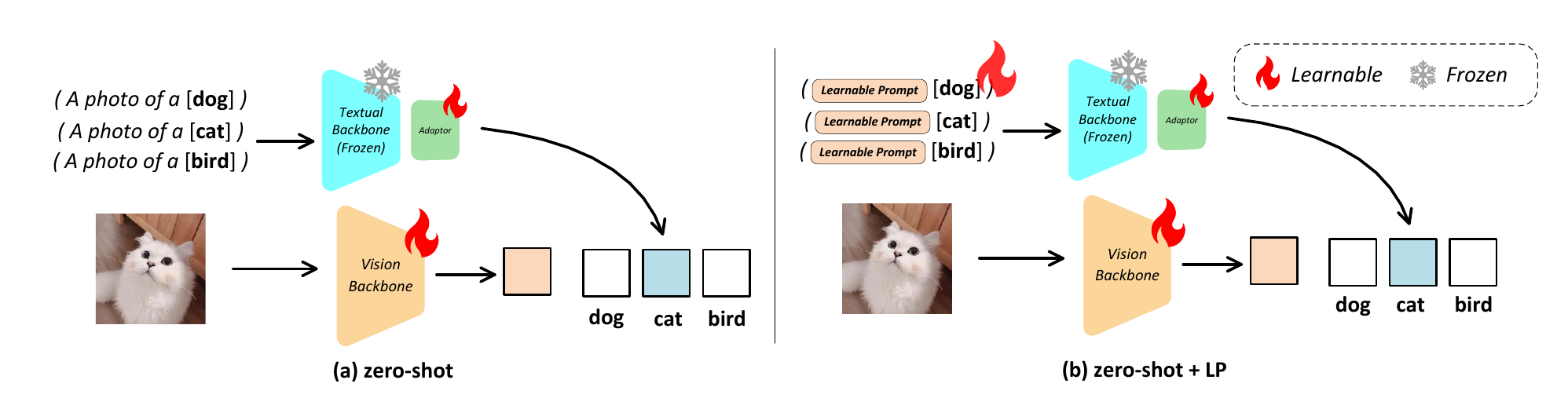}
	\caption{Illustration of the used Few-Shot Baselines with LM in Section 1, concluding the zero-shot and zero-shot+LP.}
	\label{fig:A1}
\end{figure*}

In this section, we describe the details of two simple Few-Shot Baselines with pre-trained language models (LM) we designed, as have shown in Figure 1 of Section 1, and we name them with zero-shot and zero-shot+LP, respectively. Overall, we adopt the CLIP-like \cite{radford2021learning} pre-training methodology, which sololy utilize the base dataset for training, as depicted in \Cref{fig:A1}, and we do not utilize any labels from the novel dataset following the zero-shot setting \cite{pourpanah2022review}.

\textbf{Zero-shot.} The illustration of zero-shot is shown in \Cref{fig:A1}(a). For the visual branch, we feed the images into the visual backbone to extract the visual feature. For the textual branch,  we feed the prompts into the textual backbone to extract the textual feature.  Following previous works \cite{radford2021learning,chen2023semantic}, the input prompt we chose is "A photo of a [classname]", and we employ the textual encoder of CLIP as our textual backbone. 
We also utilize an adaptor subsequent to the  textual backbone to  transform textual representations into the visual representation space. During the training stage, we aim to maximize the cosine similarity between the corresponding visual features and  textual features \cite{radford2021learning}. We only update the parameters of the visual backbone and the adaptor, while keeping the textual backbone frozen. During inference, we follow the zero-shot setting \cite{radford2021learning}, and directly predict the samples in the novel dataset without leveraging any label. 

\textbf{Zero-shot+LP.}  Building upon the zero-shot baseline, the zero-shot+LP is depicted in \Cref{fig:A1}(b). We adopt the learnable prompts instead of the pre-defined fixed prompts, while training and inference processes  remain consistent with those of the zero-shot baseline.  The parameters of the learnable prompts are updated during the training stage, and the textual backbone is still frozen. We adopt the dataset-aware prompt here as introduced in Section 4.2.  

\section{Details of Prompt Analysis}
As previously discussed in Section 5.3.2, our framework incorporates the dataset-aware learnable prompts. Additionally, we also evaluate the performances of  two alternative prompt settings: class-aware and task-aware prompts. Then we introduce the detailed configuration of these two  variants of learnable prompt.

\textbf{Class-aware Prompt.} Inspired by recent advancements in prompt Learning \cite{zhou2022learning, zhou2022conditional}, we attempt to utilize the learnable class-aware prompts, which are dynamically tailored to specific classes and conditional on the visual class prototypes.  Formally, we obtain the class-aware prompt $\mathbf{v}_{class}$ by utilizing both the dataset-aware prompt $\mathbf{v}$ and the visual prototype $\boldsymbol{p_i^{0}}$ as follows:
\begin{equation}
\mathbf{v}_{class} = \mathbf{v} + \pi (\boldsymbol{p_i^{0}})
\end{equation}
where $\pi (\ )$ denotes the class-aware Net, which is designed to generate the conditional prompt vectors for each prototype. And we employ a two-layer MLP as the class-aware Net here. Then, the obtained dataset-aware prompt $\mathbf{v}$ is fed into the textual backbone for subsequent processes.

\textbf{Task-aware Prompt.} Considering the particularity of meta-training strategy, which contains numerous few-shot learning (FSL) tasks, with each FSL task containing distinct samples and classes, we also investigate the performance of the task-aware prompt, denoted with $\mathbf{v}_{task}$. Analogous to the class-aware prompt, we obtain $\mathbf{v}_{task}$ by utilizing the dataset-aware prompt $\mathbf{v}$ and all visual prototype $\boldsymbol{p_i^{0}}$ within an few-shot learning task:
\begin{equation}
\mathbf{v}_{task} = \mathbf{v} + \pi (  \Sigma_i^K \boldsymbol{p_i^{0}})
\end{equation}
Similar to the class-aware Net, $\pi (\ )$ here denotes the task-aware Net, which aims to generate conditional prompt vectors for each FSL task, for which we also utilize a two-layer MLP.

\section{Details of Adaptor Analysis}
\vspace{-0.2cm}
\begin{figure}[ht]
	\centering
	\includegraphics[width=9cm]{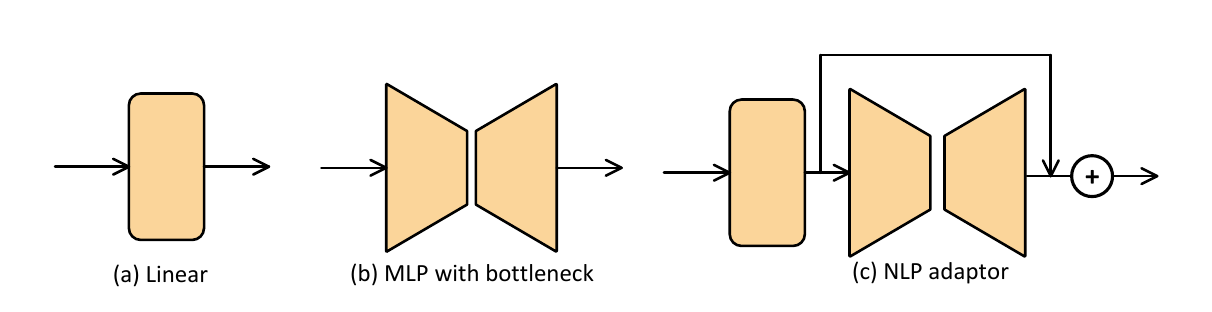}
	\caption{Illustration of three adaptors utilized in the paper}
	\label{fig:A2}
\vspace{-0.2cm}
\end{figure}

In this section, we provide details about the compared adaptors, as discussed in Section 5.3.3, containing the linear adaptor, MLP with the bottleneck, and the NLP adaptor \cite{houlsby2019parameter}.  And we visualize their structures in \Cref{fig:A2}, respectively.

\textbf{Linear adaptor.} 
The linear adaptor is prevalently utilized in previous works \cite{xing2019adaptive,chen2023semantic}. The transformed textual semantic feature $z$ can be obtained by:
\begin{align}
	z =  \boldsymbol{W_0} \ g(y^{text}_\mathbf{v}) + \boldsymbol{b_0} 
\end{align}
where $\boldsymbol{W_0}$ and $\boldsymbol{b_0}$ are the learnable parameters of the adaptor module. And the resulting $z\in \mathbb{R}^{d_v} $ has the same dimension with the visual feature $f(\boldsymbol{x_j})$.

\textbf{MLP with the bottleneck.} We adopt a two-layer Multi-Layer Perceptron (MLP) with bottleneck structure as the adaptor in our framework.  This transformation  process is expressed as:
\begin{align}
	z = \boldsymbol{W_2} \  \sigma ( \boldsymbol{W_1} \ g(y^{text}_\mathbf{v}) + \boldsymbol{b_1} ) + \boldsymbol{b_2}
\end{align}
where $\boldsymbol{W_1}$, $\boldsymbol{W_2}$, $\boldsymbol{b_1}$, and $\boldsymbol{b_2}$ are the learnable parameters. The intermediate hidden dimension we set is $\frac{d_v}{4}$.

\textbf{NLP adaptor.} Inspired by the NLP adaptor \cite{houlsby2019parameter}, we also evaluate the combination of the bottleneck structure and a linear layer with the residual connection \cite{he2016deep}. An illustration of this structure can be found in \Cref{fig:A2}(c), and formally, we have:
\begin{align}
        z' &=  \boldsymbol{W_0} \ g(y^{text}_\mathbf{v}) + \boldsymbol{b_0}   \\
	z'' &= \boldsymbol{W_2} \  \sigma ( \boldsymbol{W_1} \ z') + \boldsymbol{b_1} ) + \boldsymbol{b_2} \\
        z &= z' + z''
\end{align}
where equation (5) and (6) denote the linear structure and MLP, respectively. And the resulting $z\in \mathbb{R}^{d_v} $ in equation (7) has the same dimension with the visual feature $f(\boldsymbol{x})$. The intermediate hidden dimension of the MLP we set is $\frac{d_v}{4}$, and Equation (7) depicts the residual connection \cite{he2016deep}.

\section{Details of Fusion Mechanism}
In this section, we delve into the specifics of the differing Fusion mechanisms discussed in Section 5.3.4, particularly focusing on Concatenation and Attention mechanisms.

\textbf{Concatenation.} Recalling that we have obtained the visual feature $f(\boldsymbol{x})$ and the transformed textual semantic feature $z$, the Concatenation fusion mechanism is articulated as:
\begin{align}
        z_{f} = \boldsymbol{W_f} \ (f(\boldsymbol{x}) ||  z)
\end{align}
where $\boldsymbol{W_f}$ denotes the learnable parameters in the Fusion mechanism, and $||$ means the Concatenation operation. The generated fusion feature $z_{f}$ shares the same dimension as the visual feature $f(\boldsymbol{x})$.

\textbf{Attention.} We aim to learn the fusion weight automaticly with an attention mechanism. Formally, we have:
\begin{align}
        \alpha &= \mathbf{Att}(f(\boldsymbol{x}) ||  z) \in (0,1) \\
        z_{f} &= \alpha  z + (1- \alpha) f(\boldsymbol{x}) 
\end{align}
where $\alpha $ represents the learned fusion weight, and $\mathbf{Att}(\ )$ refers to the attention mechanism which consist of a two-layer MLP with a sigmoid activation. 

As discussed in Section 5.3.4, we empirically observe that the choice of Addition or Attention marginally affects the performance.  For the sake of simplicity, our framework adopts the additive operation.

\end{document}